\documentclass{article}

\usepackage{palatino}
\usepackage[letterpaper,margin=1in]{geometry}

\usepackage{amsthm}
\usepackage{graphicx}
\usepackage{subcaption}
\usepackage{booktabs}
\usepackage{hyperref}
\usepackage{thmtools,thm-restate}

\usepackage{amsmath}
\usepackage{bigints}

\usepackage{amssymb,amsopn,algorithm,algorithmic,float,bbm,bm,enumerate,color,multirow}

\usepackage{epsfig,graphicx}
\usepackage{comment}

\allowdisplaybreaks

\usepackage[dvipsnames]{xcolor}
\usepackage[]{color-edits}
\addauthor{ab}{blue}
\addauthor{abb}{red}
\addauthor{yz}{brown}
\addauthor{ql}{purple}
\usepackage{notation}

\title{Loss Gradient Gaussian Width based \\Generalization and Optimization Guarantees}

\author{
   Arindam Banerjee \\
   Siebel School of Computing and Data Science\\
   University of Illinois Urbana-Champaign\\
  \texttt{arindamb@illinois.edu}\\
  \and 
  Qiaobo Li \\
   Siebel School of Computing and Data Science\\
   University of Illinois Urbana-Champaign\\
  \texttt{qiaobol2@illinois.edu}\\
  \and 
  Yingxue Zhou \\
  Department of Computer Science  \\
  University of Minnesota, Twin Cities \\
  \texttt{zhou0877@umn.edu}
}

\date{}

\begin{document}

\maketitle

\begin{abstract}
Generalization and optimization guarantees on the population loss often rely on uniform convergence based analysis, typically based on the Rademacher complexity of the predictors. The rich representation power of modern models has led to concerns about this approach. In this paper, we present generalization and optimization guarantees in terms of the complexity of the gradients, as measured by the Loss Gradient Gaussian Width (LGGW). First, we introduce generalization guarantees directly in terms of the LGGW under a flexible gradient domination condition, which includes the popular PL (Polyak-Łojasiewicz) condition as a special case. 
Second, we show that sample reuse in iterative gradient descent 
does not make the empirical gradients deviate from the population gradients as long as the LGGW is small. Third, focusing on deep networks, we bound their single-sample 
LGGW in terms of the Gaussian width of the featurizer, i.e., the output of the last-but-one layer. 
To our knowledge, our generalization and optimization guarantees in terms of LGGW are the first results of its kind, 
and hold considerable promise towards quantitatively tight bounds for deep models.

\end{abstract}

\section{Introduction}
\label{sec:intro}

Machine learning theory typically characterizes generalization behavior 
in terms of uniform convergence guarantees, often in terms of Rademacher complexities of predictor function classes~\cite{kopa00,bame02,ssbd14,mort18}. However, given the rich representation power of modern deep models, it has been difficult to develop sharp uniform bounds, and concerns have been raised whether non-vacuous bounds following the uniform convergence route may be possible~\cite{nagarajan_uniform_2019,negrea_defense_2019,kast09,kast12,bartlett_spectrally-normalized_2017,golowich_size-independent_2017}. On the other hand, the empirical successes of deep learning has highlighted the importance of gradients, which play a critical role in learning over-parameterized models with deep representations~\cite{bottou_optimization_2016,gobc16,lecun_deep_2015}. The motivation behind this paper is to investigate generalization and optimization guarantees based on uniform convergence analysis of gradients, rather than the predictors. 

Empirical work in recent years has demonstrated that gradients of overparameterized deep learning models are typically ``simple,'' e.g., only a small fraction of entries have large values \cite{papyan_full_2018,papyan_measurements_2019, lgzcb20, wu2022dissecting, xie2022rethinking}, gradients typically lie in a low-dimensional space \cite{gur-ari_gradient_2018, ghorbani_investigation_2019,rothchild2020fetchsgd,ivkin2019communication}, 
using top-$k$ components with suitable bias corrections work well in practice~\cite{rothchild2020fetchsgd,ivkin2019communication}, 
etc. In this paper, we view such simplicity in terms of a generic chaining based ``covering" of the gradients, which can be represented as the Loss Gradient Gaussian Width (LGGW)~\cite{tala14,vershynin_high-dimensional_2018}. Based on the empirical evidence, we posit that loss gradients of modern machine learning models have small LGGW.
Our main technical results in the paper present generalization and optimization guarantees 
in terms of the LGGW. 
While our main technical results hold irrespective of whether the LGGW is small or not, similar to their classical Rademacher complexity counterparts, the results imply sharper bounds if the LGGW is small. Further, we consider deep learning models, specifically Feed-Forward Networks (FFNs) and Residual Networks (ResNets), and show that under standard assumptions, the LGGW of these models can be expressed in terms of the Gaussian width of the featurizer, i.e., the output of the last-but-one layer. 





In Section~\ref{sec:gen}, we present generalization bounds in terms of the LGGW, based on a generic chaining argument over the gradients. Our analysis needs a mechanism to transition from losses to gradients of losses, and we assume the population loss function satisfies a flexible gradient domination (GD) condition~\cite{FSS}, which includes the PL (Polyak-Łojasiewicz) condition~\cite{Polyak} used in modern deep learning~\cite{liu_toward_2020,CL-LZ-MB:20,liu_loss_2021} as a simple special case. 
Then, based on recent advances in uniform convergence of vector-valued functions~\cite{maur16,mapo16,FSS}, 
one can readily get generalization bounds based on the vector Rademacher complexity of gradients. Our main new result on generalization is to show that such vector Rademacher complexity can be bounded in terms of the LGGW, which makes the bound explicitly depend on the geometry of gradients. 
Our proof is based on generic chaining (GC)~\cite{tala14}, where we first bound the vector Rademacher complexity using a hierarchical covering argument, and then bound the hierarchical covering in terms of the LGGW using the majorizing measure theorem~\cite{tala14}. To our knowledge, this is the first generalization bound in terms of Gaussian width of loss gradients, and the results imply that models with small LGGW will generalize well. 

In Section~\ref{sec:sgd}, we present optimization guarantees based on the LGGW. One  big disconnect between the theory and practice of (stochastic) gradient descent (GD) is that the theory assumes fresh samples in each step whereas in practice one reuses samples. If one just focuses on the finite sum empirical risk minimization (ERM), sample reuse is not an issue. However, the goal of such optimization is to reduce the population loss, and sample reuse derails the standard analysis. We show that even with sample reuse, the discrepancy between the sample average and the population gradient can be bounded by the LGGW along with only a logarithmic dependence on the number of steps of optimization, i.e., the number of times the samples has been reused, and also a logarithmic dependence on the ambient dimensionality. Thus, if the LGGW is small, the sample average gradients stay close to the population gradients, even under sample reuse. We also use these results to establish population convergence results for GD with sample reuse. 

In Section~\ref{sec:rad_grad}, we present the first results on bounding LGGW of deep learning gradients. 
We consider formal models for FFNs and ResNets widely studied in the deep learning theory literature~\cite{allen-zhu_convergence_2019,du_gradient_2019,arora_exact_2019,liu_loss_2021,banerjee_rsc23} and show that their single-sample LGGW can be bounded by the Gaussian width of the featurizer, i.e., the output of the last-but-one layer. We also demonstrate benefits of architecture choices, e.g., hidden layers being wider, last-but-one layer being narrower, which yields smaller single-sample LGGW.




Notation: $c_1, c_2, c_3$, etc., denote constants, and they may mean different constants at different places. 
We use $\tilde{O}$ to hide poly-log terms.

\section{Gaussian Width based Generalization Bounds}
\label{sec:gen}

In this section, we formalize the notion of  Loss Gradient Gaussian Width (LGGW) \cite{vershynin_high-dimensional_2018, tala14}, and as the main result, establish generalization bounds in terms of LGGW. For parameters $\theta \in \Theta \subset \R^p$ and data points $z \in \cZ$ where $z=(\x,y), \x \in \R^d, y \in \R$, for any suitable loss $\ell(\theta;z): \Theta \times \cZ \mapsto \R_+$, the gradients
of interest are $\xi(\theta;z) = \nabla_{\theta} \ell(\theta;z) \in \R^p$. 
Given a dataset of $n$ samples, we start by defining the sets of all $n$-tuple of individual gradients and scaled stacked gradients: 
\begin{defn}[{\bf Gradient Sets}]\label{defn:gradset}
Given $n$~samples $z^{(n)} = \{z_1,\ldots,z_n\} \in \cZ^{(n)}$, for a suitable choice of parameter set $\Theta$ and loss $\ell : \Theta \times \cZ \mapsto \R_+$, let the set of all possible $n$-tuple of individual loss gradients and scaled stacked gradients be respectively denoted as:
\begin{align}
\hat{\Xi}^{(n)}(\Theta) & := \left\{ \hat{\xi}^{(n)} = \{\xi_i, i \in [n]\} \mid 
\exists \theta \in \Theta ~\text{s.t.}~ \xi_i = \nabla \ell(\theta;z_i)  \right\}~, \label{eq:hatxin} \\
\hat{\Xi}_{n}(\Theta) & := \left\{ \hat{\xi}_n \in \R^{np} \mid \exists \theta \in \Theta~~~\text{s.t.}~~\hat{\xi}_n = \frac{1}{\sqrt{n}} \big[\nabla \ell(\theta;z_i)\big]_{i=1}^n \right\}~, \label{eq:hatxi} 
\end{align}
where $\big[\nabla \ell(\theta;z_i)\big]_{i=1}^n$ denotes the $n$ $p$-dimensional gradients stacked to create $\hat{\xi}_n \in \R^{np}$.
\end{defn}
The scaling of $\frac{1}{\sqrt{n}}$ in the stacked gradient vector $\hat{\xi}_n$ is to maintain the $L_2$-norm at the same scale as individual gradient components, i.e., with $\| \nabla \ell(\theta;z_i) \|_2 = O(1)$, we will have $\| \hat{\xi}_n \|_2 = O(1)$. 

For convenience, we will denote the set of $n$-tuples of gradients in \eqref{eq:hatxin} as $\hat{\Xi}^{(n)}$ and the scaled stacked gradients in \eqref{eq:hatxi} as $\hat{\Xi}_n$. Note that the sets $\hat{\Xi}^{(n)}$ and $\hat{\Xi}_n$ are conditioned on $z^{(n)}$, and the results we will present will hold for any $z^{(n)} \in \cZ^n$, i.e., any $n$ samples under suitable regularity conditions (see Section~\ref{ssec:ffn}). 
Next, we define the Loss Gradient Gaussian Width (LGGW):
\begin{defn}[{\bf Loss Gradient Gaussian Width}]\label{defn:gaussw}
Given the scaled stacked gradient set $\hat{\Xi}_n$ as in \eqref{eq:hatxi} in Definition~\ref{defn:gradset}, the Loss Gradient Gaussian Width (LGGW) is defined as:
\begin{align}
w(\hat{\Xi}_n) := \E_{\g \sim \cN(0, \I_{np \times np})}\left[\sup_{\hat{\xi}_n \in \hat{\Xi}_n}~\langle \hat{\xi}_n, \g \rangle \right]~.
\end{align}
\end{defn}
We refer the reader to the book by Vershynin~\cite{vershynin_high-dimensional_2018} for a technical introduction to Gaussian width and its properties. At a high level, 
Gaussian width measures how much elements of the set aligns with random Gaussian vectors, capturing the complexity of the set. As a special case of generic chaining (GC), Gaussian width estimates this alignment using expected supremum over Gaussian processes indexed on the set~\cite{tala14}. The GC perspective forms the basis of our technical development and we elaborate on this perspective further in the sequel. 


\subsection{Bounds based on Vector Rademacher Complexity}


Our approach to generalization bounds is based on vector Rademacher complexities. Unlike the traditional approach which focus on bounds based on Rademacher complexities of the set of predictors $f(\theta;x)$ \cite{kopa00,bame02,ssbd14,mort18}, our work will leverage
generalization bounds based on  vector Rademacher complexities \cite{maur16,mapo16,FSS} of the set of loss gradients $\xi(\theta,z) = \nabla \ell(\theta;z)$. In particular, following~\cite{FSS}, we consider {\em Normed Empirical Rademacher Complexity} (NERC), which is the extension of standard Rademacher complexity to sets of vectors or vector-valued functions.
\begin{defn}\label{defn:nerc}
In the setting of Definition~\ref{defn:gradset}, for the set $\hat{\Xi}^{(n)}$ of $n$-tuples of individual gradients, the Normed Empirical Rademacher Complexity (NERC) is defined as
\begin{align}
\hat{R}_n(\hat{\Xi}^{(n)}) & :=  \frac{1}{n} \E_{\eps^{(n)}}\left[ \sup_{ \hat{\xi}^{(n)} \in \hat{\Xi}^{(n)}} \left\| 
\sum_{i=1}^n \eps_i \xi_i \right\| ~  \right] ~,  \label{eq:nerc} 
\end{align}
where $\eps^{(n)}$ are a set of $n$ i.i.d.~Rademacher random variables, i.e., $P(\eps_i=+1)=P(\eps_i=-1)=\frac{1}{2}$, and $\hat{\Xi}^{(n)}$ is the set of $n$-tuples of gradients as in \eqref
{eq:hatxin}.
\end{defn}

For \eqref{eq:nerc}, we present results for the $L_2$ norm $\| \cdot \|_2$, but the analyses tools and results have counterparts for general norms~\cite{leta91} and such results will be in terms of $\gamma$-functions in generic chaining~\cite{tala14}. 

\begin{figure*}[t]
    \centering
    \begin{subfigure}[b]{0.28\textwidth}
        \includegraphics[width=\textwidth]{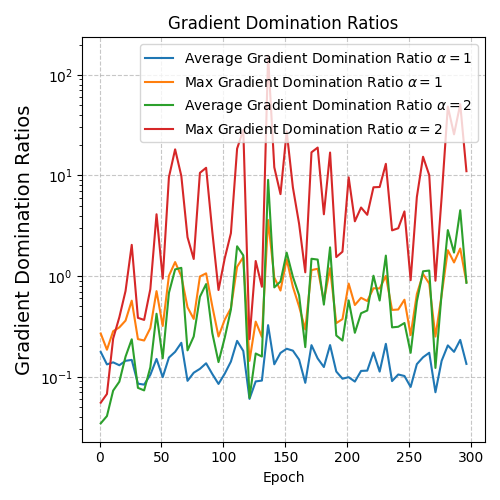}
        \caption{ResNet18 on CIFAR-10}
        \label{fig:resnet18_cifar10}
    \end{subfigure}
    \hfill
    \begin{subfigure}[b]{0.28\textwidth}
        \includegraphics[width=\textwidth]{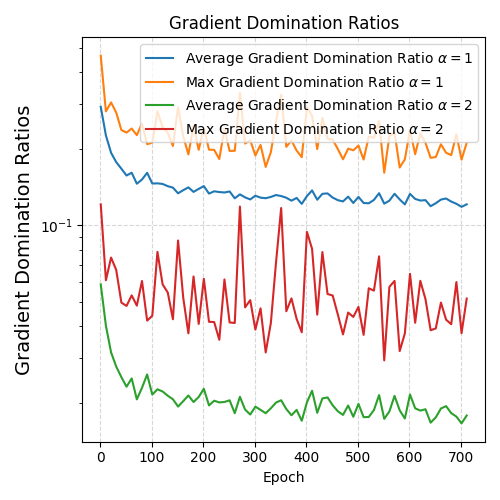}
        \caption{FFN on CIFAR-10}
        \label{fig:ffn_cifar10}
    \end{subfigure}
    \hfill
    \begin{subfigure}[b]{0.28\textwidth}
        \includegraphics[width=\textwidth]{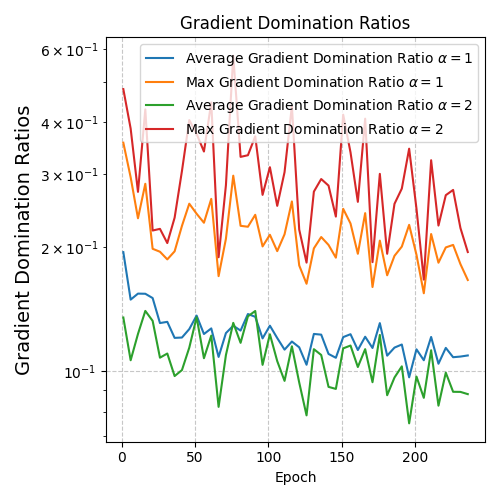}
        \caption{CNN on Fashion-MNIST}
        \label{fig:cnn_fashion_mnist}
    \end{subfigure}
    \caption{Gradient Domination (GD) Ratio for $\alpha=1, 2$ for (a) ResNet18 on CIFAR-10, (b) FFN on CIFAR-10, and (c) CNN on Fashion-MNIST. Each experiment is repeated 5 times and we report the average and the maximum ratio. The results show that GD holds for $\alpha=1$ with a small constant $\bar{c}_{1} < 5$ for all models, and for $\alpha=2$, with $\bar{c}_{2} < 1$ for some models, but needs large constants $\bar{c}_{2} > 100$ for for ResNet18.}
    \label{fig:gradient_domination_ratio}
\end{figure*}

Given a loss function $\ell$ and a distribution $\cD$ over examples $z \in \cZ$, we consider the minimization of the \emph{population risk} function, that is,
 \begin{equation} \label{eq:problem_prm}
 \min_\theta \cL_{\cD}(\theta) := \E_{z\sim \cD}\left[ \ell(\theta; z) \right]~. 
 \end{equation}
The learner does not observe the distribution $\cD$ and typically minimizes the \emph{empirical risk} 
 \begin{equation}
 \hat \cL_n(\theta) := \frac{1}{n} \sum_{i=1}^n \ell(\theta; z_i)~,
 \end{equation}
where $z^{(n)} \sim \cD^n$ are i.i.d.~draws. While the loss functions are usually non-convex in modern machine learning, including deep learning, they usually satisfy a form of the \emph{gradient domination} (GD) condition~\cite{FSS}, which includes the popular Polyak-Łojasiewicz (PL) condition as a simple special case~\cite{Polyak,liu_toward_2020,CL-LZ-MB:20,liu_loss_2021}. Further, we assume the gradients to be bounded. 
\begin{asmp}[{\bf Population Risk: Gradient Domination}]
\label{asmp:gd_cond}
The population risk $\cL_\cD$ satisfies the $(\alpha, \bar{c}_{\alpha})$-GD condition with respect to a norm $\|\cdot \|$, if there are constants $\bar{c}_{\alpha} > 0$, $\alpha \in [1,2]$, such that $\forall \theta \in \Theta$ and $\theta^{\star} \in \argmin _{\theta \in \Theta} \cL_{\mathcal{D}}(w)$, we have 
 \begin{align}
 \cL_{\mathcal{D}}(\theta)-\cL_{\mathcal{D}}\left(\theta^{\star}\right) \leq \bar{c}_{\alpha} \left\|\nabla \cL_{\mathcal{D}}(\theta)\right\|^{\alpha}~.
 \label{eq:plcond}
 \end{align}
\end{asmp}
\begin{asmp}[{\bf Bounded Gradients}]\label{asmp:gradbnd}
The sample gradients $\xi(\theta;z) = \nabla \ell(\theta;z)$ are bounded, i.e., $\|\xi(\theta;z)\|_2 =O(1), \theta \in \Theta, z \in \cZ$. 
\end{asmp}
\begin{remark}
To understand whether the $(\alpha,\bar{c}_{\alpha})$-GD condition in Assumption~\ref{asmp:gd_cond} holds in deep learning models, 
in Figure~\ref{fig:gradient_domination_ratio}, we plot the gradient domination ratio $\frac{\cL_{\mathcal{D}}(\theta)}{\left\|\nabla \cL_{\mathcal{D}}(\theta)\right\|_2^{\alpha}}$ for $\alpha=1, 2$ for three standard deep learning models: ResNet, FFN, and CNN.
The plots demonstrate that the condition indeed empirically holds in the optimization trajectory for $\alpha=1$ with a small constant $\bar{c}_{1} < 5$ for all models. For $\alpha=2$, i.e., the PL condition \cite{Polyak,liu_toward_2020,CL-LZ-MB:20,liu_loss_2021}, the condition holds
with $\bar{c}_{2} < 1$ for some models, but needs large constants $\bar{c}_{2} > 100$ for for ResNet18 on CIFAR-10.  
\qed 
\end{remark}
\begin{remark}
The GD condition is more flexible than the PL condition as it allows for a more relaxed relationship between function values and the gradient norm. While both conditions may fail in landscapes with multiple valleys, starting optimization from a pre-trained model often positions the model within a single valley, where the GD condition is likely to hold. 
\qed
\end{remark}

We start with a generalization bound from~\cite{FSS} for losses satisfying gradient domination. 
\begin{restatable}{prop}{genbnd}
\label{prop:genbnd}
Under Assumptions~\ref{asmp:gd_cond} and \ref{asmp:gradbnd}, with $\theta^{\star} \in \argmin _{\theta \in \Theta} \cL_{\mathcal{D}}(\theta)$ denoting any population loss minimizer and $\hat{R}_n(\hat{\Xi}^{(n)})$ as in Definition~\ref{defn:nerc}, for any $\delta > 0$, with probability at least $1-\delta$ over the draw of the samples $z^{(n)} \sim \cD^n$, for any $\hat{\theta} \in \Theta$ we have 
\begin{equation}
\cL_{\cD}(\hat{\theta}) - \cL_{\cD}(\theta^*)   \leq 2 \bar{c}_{\alpha} \bigg( \| \nabla \hat{\cL}_n(\hat{\theta}) \|_2^{\alpha}+ 2\left( 4 \hat{R}_n(\hat{\Xi}^{(n)}) + c_3 \frac{\log \frac{1}{\delta}}{n} \right)^{\alpha} \bigg) + c_4 \left( \frac{\log \frac{1}{\delta}}{n} \right)^{\frac{\alpha}{2}}  ~.
\end{equation}
\end{restatable}
For such a bound to be useful, the NERC $\hat{R}_n(\Xi^{(n)})$ needs to be small, e.g., $\hat{R}_n(\hat{\Xi}^{(n)}) = O(\frac{1}{\sqrt{n}})$. \cite{FSS} analyzed the NERC for loss functions of the form $\ell(\theta, (x, y)) =  h(\psi_{\theta}(x))$, where $\psi_{\theta}(x) = \langle \theta, x \rangle$ and $h$ is Lipschitz. For this case, it suffices to bound the NERC of the gradient of linear predictor $\psi_{\theta}(x) = \theta^T x$. Since  $\nabla \psi_{\theta}(x) = x$ is independent of $\theta$, as long as each $\|x_i\|$ is bounded (Assumption~\ref{asmp:gradbnd}), we have
$\frac{1}{n} \E_{\eps^{(n)}} \left[ \sup_{\theta \in \Theta} \left\| \sum_{i=1}^n \eps_i \nabla \phi(\theta, x_i) \right\| \right] = \frac{1}{n} \E_{\eps^{(n)}} \left[ \left\| \sum_{i=1}^n \eps_i x_i \right\| \right] = \frac{c}{\sqrt{n}}$,
by Khintchine's inequality~\cite{leta91,vers12} for $L_2$ norm and under suitable assumptions on certain other norms~\cite{kast09,kast12,FSS}. The key simplification here is that the $\sup_{\theta \in \Theta}$ drops out since the gradient does not depend on $\theta$. However, simplification is only possible for linear predictors. 
Our analysis considers the general case where the $\sup_{\theta}$ does not drop out and has to be handled explicitly.

\subsection{Main Result: Gaussian Width Bound}

For our main result, we consider whether the NERC $\hat{R}_n(\hat{\Xi}^{(n)})$ can somehow be related to the Gaussian width $w(\hat{\Xi}_n)$ as in \eqref{eq:hatxi}, which arguably captures the geometry of the gradients more directly. 

\begin{restatable}{theo}{rcgc}
\label{theo:rcgc}
Based on Definitions~\ref{defn:gradset}, \ref{defn:gaussw}, and \ref{defn:nerc}, with $\varepsilon^{(n)}$ denoting $n$ i.i.d.~Rademacher variables, conditioned on any $z^{(n)} \in \cZ^{(n)}$, for any $u>0$, with probability at least $(1-c_0 \exp(-u^2/2))$ over the randomness of  $\varepsilon^{(n)}$, we have
\begin{align} \label{eq:gc_bound}
\sup_{\hat{\xi}^{(n)} \in \hat{\Xi}^{(n)}} \frac{1}{\sqrt{n}} \left\| \sum_{i=1}^n \eps_i \xi_i \right\|_2 \leq c_5 (1 + u)w(\hat{\Xi}_n)~.
\end{align}
As a result, the normed empirical Rademacher complexity (NERC) satisfies 
\begin{equation} \label{eq:rc_bo}
     \hat{R}_n(\hat{\Xi}^{(n)}) = \frac{1}{n} \E_{\eps^{(n)}} \left[ \sup_{\hat{\xi}^{(n)} \in \hat{\Xi}^{(n)}} \left\| \sum_{i=1}^n \eps_i \xi_i \right\|_2 \right] \leq  \frac{c_{6}w(\hat{\Xi}_n)}{\sqrt{n}}~.
\end{equation}
\end{restatable}
One can plug-in the Theorem~\ref{theo:rcgc} to Proposition~\ref{prop:genbnd} to get the desired generalization bound. For example, with Assumption~\ref{asmp:gd_cond} holding for $\alpha=1$ (Figure~\ref{fig:gradient_domination_ratio}), with probability $(1-\delta)$, for any $\hat{\theta} \in \Theta$ we have
\begin{equation}
\begin{split}
&\cL_{\cD}(\hat{\theta}) - \cL_{\cD}(\theta^*)   \leq 2  \bar{c}_1 \left( \| \nabla \hat{\cL}_n(\hat{\theta}) \|_2  + 2\left( \frac{4 c_6 w(\hat{\Xi}_n)}{\sqrt{n}}  + c_3 \frac{\log \frac{1}{\delta}}{n} \right) \right) + c_4 \sqrt{\frac{\log \frac{1}{\delta}}{n}} ~.
\end{split}
\label{eq:genbndex}
\end{equation}

\begin{remark}
A technically curious aspect of the result in \eqref{eq:rc_bo} is that the expectation in $\hat{R}_n(\hat{\Xi}^{(n)})$ is from $n$ Rademacher variables over samples whereas that in $w(\hat{\Xi}_n)$ is from $p$ normal variables over gradient components. The shift from samples to gradient components makes the bound geometric. We elaborate on such geometry in the next remark.
\qed 
\end{remark}
\begin{remark}
Since the bound is with respect to the Gaussian width of the set of scaled stacked gradients $\hat{\xi}_n = \frac{1}{\sqrt{n}} \big[\nabla \ell(\theta;z_i)\big]_{i=1}^n$, it would depend on both the alignment of the $n$ sample gradients $\xi_i = \nabla \ell(\theta;z_i)$ as well as the width of the set of individual sample gradients $\Xi_i = \left\{\hat{\xi}_i \in\R^{p}\mid \exists \theta \in \Theta, ~\text{s.t.}~~\hat{\xi}_i =\nabla\ell\left(\theta;z_i\right)\right\}$. 
Depending on the alignment over the $n$ sample gradients, the width can range from $O(1)$ to $O(\sqrt{n})$, similar to Rademacher complexity~\cite{ssbd14,bame02}. The nature of alignment here will be based on similarity of the hierarchical covering of individual gradient sets, with similar coverings, e.g., similar distortions at each level of the hierarchy, yielding small width.
For the width of individual sample gradient sets, we present bounds for two commonly used networks: FeedForward Networks and Residual Networks in Section \ref{sec:rad_grad}. \qed 
\end{remark}
\begin{remark}
It is important to understand that the bound in \eqref{eq:genbndex} does not depend on the algorithm used to reach $\hat{\theta} \in \Theta$. Thus, one can use variants of stochastic gradient descent, adaptive optimization, or sharpness aware minimization \cite{zaheer_adaptive_2018, reddi_convergence_2018,foret2020sharpness} to reach $\hat{\theta}$, and the bound would still hold. \qed 
\end{remark}

\subsection{Proof sketch of Main Result} 

Our analysis is based on generic chaining (GC) and needs the following definition~\cite{tala14}:
\begin{defn}
	For a metric space $(T,d)$, an admissible sequence of $T$ is a collection of subsets of $T$, $\Gamma = \{T_r : r \geq 0\}$,
	with $|T_0| = 1$ and $|T_r| \leq 2^{2^r}$ for all $r \geq 1$. For $\beta \geq 1$, the $\gamma_{\beta}$ functional is defined by  
	\begin{equation}
	\gamma_{\beta}(T,d) := \inf_\Gamma \sup_{t \in T} \sum_{r=0}^{\infty} 2^{r/\beta} d(t,T_r)~,
	\end{equation}
	where the infimum is over all admissible sequences $\Gamma$ of $T$.
\end{defn}
\begin{remark}
There has been substantial developments on GC over the past two decades~\cite{tala14,talagrand_majorizing_1996}. GC lets one develop sharp upper bounds to suprema of stochastic processes indexed in a set with a metric structure in terms of $\gamma_\beta$ functions. Typically, the bounds are in terms of $\gamma_2$ functions for sub-Gaussian processes and a mix of $\gamma_1$ and $\gamma_2$ functions for sub-exponential processes. \qed 
\end{remark}

To get to and understand the proof of Theorem~\ref{theo:rcgc}, we first recap the standard GC argument. For any suitable stochastic process under consideration, say $\{X_\xi\}$, the key step is to find a suitable pseudo-metric $d(\xi_1,\xi_2)$ that satisfies {\em increment condition} (IC)~\cite{tala14}: for any $u > 0$
\begin{equation}
    \P(|X_{\xi_1} - X_{\xi_2}| \geq ud(\xi_1,\xi_2) ) \leq 2 \exp(-u^2/2)~.
    \label{eq:gcic2}
\end{equation}
Then, under some simple assumptions, e.g., $X_\xi = 0$ for some $\xi \in \Xi$, GC shows that $\E[\sup_{\xi \in \Xi} X_\xi] \leq c \gamma_2(\Xi,d)$, for some constant $c$, and there is a corresponding high probability version of the result~\cite{tala14}. The analysis is typically done using the canonical distance
\begin{align}
d(\xi_1,\xi_2) := (\E[ (X_{\xi_1} - X_{\xi_2})^2])^{1/2}~,
\label{eq:canon2}
\end{align}
though one can use other (not canonical) distance $\bar{d}(\xi_1,\xi_2)$ which satisfies the increment condition \eqref{eq:gcic2}.

At a high level, our proof has three parts:
\begin{itemize}
\item Since the NERC $\hat{R}_n(\hat{\Xi}^{(n)})$ is defined in terms of the set of $n$-tuples $\hat{\Xi}^{(n)}$, we start with a suitable (non canonical) distance $\bar{d}^{(n)}(\hat{\xi}^{(n)}_1,\hat{\xi}^{(n)}_2)$ over $n$-tuples in $\hat{\Xi}^{(n)}$ and establish a version of Theorem~\ref{theo:rcgc} in terms of $\gamma_2(\hat{\Xi}^{(n)},\bar{d}^{(n)})$, showing $\hat{R}_n(\hat{\Xi}^{(n)}) \leq c_{2}\frac{\gamma_2(\hat{\Xi}^{(n)},\bar{d}^{(n)})}{\sqrt{n}}$. 
\item For the set $\hat{\Xi}_n$ of stacked gradients $\hat{\xi}_n$, we construct a Gaussian process $X_{\hat{\xi}_n} = \langle \hat{\xi}_n, \g\rangle$ where $\g \sim \cN(0,\I_{np \times np})$. 
We note a simple one-to-one correspondence between the set of stacked gradients $\hat{\Xi}_n$ and the set of $n$-tuples $\hat{\Xi}^{(n)}$ since $\hat{\xi}_n = \frac{1}{\sqrt{n}}\left[\xi_i\right]_{i=1}^{n}$. Further, 
we show that the canonical distance $d(X_{\hat{\xi}_{1,n}}, X_{\hat{\xi}_{2,n}})$ as in \eqref{eq:canon2} from the Gaussian process on $\hat{\Xi}_n$ satisfies $d(X_{\hat{\xi}_{1,n}}, X_{\hat{\xi}_{2,n}}) = \bar{d}^{(n)}(\hat{\xi}^{(n)}_1,\hat{\xi}^{(n)}_2)$, the non-canonical distance on $\hat{\Xi}^{(n)}$. Based on the set correspondence and exact same distance measure, we have $\gamma_2(\hat{\Xi}_n, d) = \gamma_2(\hat{\Xi}^{(n)}, \bar{d}^{(n)})$, so that $\hat{R}_n(\hat{\Xi}^{(n)}) \leq c_{2}\frac{\gamma_2(\hat{\Xi}_n,d)}{\sqrt{n}}$. 
\item The final step of the analysis is to show that $\gamma_2(\hat{\Xi}_n,d) \leq c_{4} w(\hat{\Xi}_n)$. The result follows directly from the Majorizing Measure Theorem~\cite{tala14} since we are working with a Gaussian process indexed on $\hat{\Xi}_n$ and using the canonical distance $d$. 
\end{itemize}

We provide additional details and associated lemmas 
for the first part of the analysis. Consider the stochastic process $X_{\hat{\xi}^{(n)}} := \frac{1}{\sqrt{n}} \left\| \sum_{i=1}^n \eps_i \xi_i \right\|_2$, by triangle inequality, we have
\begin{align}
\left|X_{\hat{\xi}^{(n)}_1} - X_{\hat{\xi}^{(n)}_2}\right| \leq \frac{1}{\sqrt{n}} \left\| \sum_{i=1}^n \eps_i (\xi_{i,1} - \xi_{i,2}) \right\|_2~. 
\label{eq:jencan}
\end{align}
Thus, for a suitable (non canonical) metric $\bar{d}^{(n)}(\hat{\xi}^{(n)}_1,\hat{\xi}^{(n)}_2)$, it suffices to show an increment condition (IC) of the form: for any $u>0$, with probability at least $1-2  \exp(-u^2/2)$ over the randomness of $\eps^{(n)}$, we have 
\begin{equation}
\frac{1}{\sqrt{n}} \left\| \sum_{i=1}^n \eps_i (\xi_{i,1} - \xi_{i,2}) \right\|_2 \leq u \bar{d}^{(n)}(\hat{\xi}^{(n)}_1,\hat{\xi}^{(n)}_2) ~.
\label{eq:vh1}
\end{equation}
The in-expectation version of such a result is usually easy to establish using Jensen's inequality, e.g., see (26.16) in \cite{ssbd14}. The high probability version can be more tricky. We focus on a variant of the increment condition which allows for a constant shift, i.e., right hand side of the form $(\mu+u) \bar{d}^{(n)}(\hat{\xi}^{(n)}_1,\hat{\xi}^{(n)}_2)$ where $\mu=O(1)$. First, we consider the non-cannnical distance 
\begin{align}
\bar{d}^{(n)}(\hat{\xi}^{(n)}_1,\hat{\xi}^{(n)}_2) := \left( \frac{1}{n} \sum_{i=1}^n \| \xi_{i,1} - \xi_{i,2} \|_2^2 \right)^{1/2} 
\label{eq:ncdist}
\end{align}
which is greater or equal to the canonical distance as in \eqref{eq:canon2}, due to triangle inequality in \eqref{eq:jencan}.
With our choice of distance $\bar{d}^{(n)}$, note that $\v_i = \frac{\xi_{i,1} - \xi_{i,2}}{\bar{d}^{(n)}(\hat{\xi}^{(n)}_1,\hat{\xi}^{(n)}_2)}$ satisfies $\sum_{i=1}^n \|\v_i\|^2_2 = n$. Then, the following result is effectively a {\em shifted increment condition} (SIC), a shifted variant of \eqref{eq:vh1} with a constant shift of $\mu \leq 1$:
\begin{restatable}{lemm}{vechf}
Let $\v_i \in \R^p, i=1,\ldots,n$ be a set of vectors such that $\sum_{i=1}^n \| \v_i \|^2_2 \leq n$. Let $\eps_i$ be a set of i.i.d.~Rademacher random variables. 
Then, for any $u > 0$, 
\begin{equation}
\P_{\eps^{(n)}} \left( \left| \left\| \frac{1}{\sqrt{n}} \sum_{i=1}^n \eps_i \v_i \right\|_2 - \mu \right| \geq u \right) 
    \leq 2\exp \left(- \frac{u^2}{2}\right)~,
\end{equation}
where $\mu \leq 1$ is a positive constant.
\label{thm:vechf}
\end{restatable}
Next, we show that such constant shifts can be gracefully handled by GC:
\begin{restatable}{lemm}{shiftgc}
Consider a stochastic process $\{ X_{\hat{\xi}^{(n)}} \}, \hat{\xi}^{(n)} \in \hat{\Xi}^{(n)}$ which satisfies: for any $u > 0$, with probability at least $1-c_1 \exp(-u^2/2)$, we have  
\begin{equation} \label{eq:ic}
 \left|X_{\hat{\xi}^{(n)}_1} - X_{\hat{\xi}^{(n)}_2}\right| \leq  c_0(\mu + u) \bar{d}^{(n)}\left(\hat{\xi}^{(n)}_1,\hat{\xi}^{(n)}_2\right) ~.
\end{equation}
Further, assume that $X_{\hat{\xi}^{(n)}_0}=0$ for some $\hat{\xi}^{(n)}_0 \in \hat{\Xi}^{(n)}$. Then, 
for any $u > 0$, with probability at least $1-c_3 \exp(-u^2/2)$, we have 
\begin{equation} \label{eq:ic_gc}
 \sup_{\hat{\xi}^{(n)} \in \hat{\Xi}^{(n)}} \left|X_{\hat{\xi}^{(n)}} \right|  \leq c_2(\mu + u) \gamma_2\left(\hat{\Xi}^{(n)},\bar{d}^{(n)} \right) ~.
\end{equation}
\label{thm:shiftgc}
\end{restatable}
\begin{remark}
The $X_{\hat{\xi}^{(n)}_0}=0$ is easy to satisfy as long as the gradients are all zero for some $\theta_0$, e.g., minima of the loss, interpolation condition, etc. That condition is not necessary, and a more general result can be established, e.g., see \cite{tala14}[Theorem 2.4.12]. \qed 
\end{remark} 

The SIC in Lemma~\ref{thm:vechf} and the shifted variant of GC in Lemma~\ref{thm:shiftgc} can now be used to establish the first part of the proof of Theorem~\ref{theo:rcgc}. 
The technicalities for the other parts are more straightforward and are directly handled in the proof of Theorem~\ref{theo:rcgc}. All proofs are in Appendix~\ref{app:gen}.

\section{Gradient Descent with Sample Reuse}
\label{sec:sgd}

Gradient descent (GD) updates parameters following 
$\theta_{t+1} = \theta_t - \eta_t \nabla \hat{\cL}_n(\theta_t)$. If the goal is to make sure GD decreases the population loss $\cL(\theta)$, then one typically assumes that fresh samples are used in every iteration $t$ to compute $\hat{\cL}_n(\theta_t)$ and corresponding $\nabla \hat{\cL}_n(\theta_t)$, making sure such estimates are unbiased estimates of the population loss gradient. 
In practice, however, one reuses samples from the fixed training set for GD and also for mini-batch stochastic gradient descent. Such sample reuse violates the fresh sample assumption, and 
in turn cannot guarantee the unbiasedness of the gradient estimates~\cite{chba18,stup19}. While one can still guarantee decrease of the finite sum empirical loss from the ERM perspective~\cite{shalev-shwartz_learnability_2010,ghadimi_optimal_2012,nemirovski_robust_2009}  with GD, it is unclear if the population loss should decrease with such updates. Thus, the theory and practice of (stochastic) GD and variants have diverged, since the theory continues to assume fresh samples~\cite{zaheer_adaptive_2018,ghadimi_optimal_2012} whereas in practice one does multiple passes (epochs) over the fixed dataset~\cite{haochen2019random, safran2020good}.
If ERM optimization is done via sample reuse,  one needs to then use uniform convergence (UC) to get population results~\cite{shalev-shwartz_learnability_2010,ssbd14}. While this approach is well established, UC analysis over large function classes cannot quite take advantage of the geometry of gradients \cite{mei2018landscape,amir2022thinking, davis2022graphical}. 

In this section, we show that empirical estimates of the gradient with sample reuse stay close to the population gradients as long as the LGGW is small. 
%
The consequence of sample reuse  is that $\theta_t$ was determined using the samples $\{z_i, i \in [n]\}$ in prior iterations and the gradient estimate $\nabla \hat{\cL}_n(\theta_{t})=\frac{1}{n} \sum_{i=1}^n \ell(\theta_{t},z_i)$ also uses the same samples. Since $\theta_t$ is not independent of the samples, basic concentration bounds such as Hoeffding or McDiarmid are not directly applicable to $\hat{\cL}_n(\theta_{t})$. The sample reuse here is a specific case of adaptive data analysis, and 
%
has been acknowledged and studied in the adaptive data analysis literature~\cite{DworkFHPRR15,SteinkeU16}.

We start by consider the joint event of interest under such sequential sample reuse. The initialization parameter $\theta_0$ for gradient descent is assumed to be independent of the samples, e.g., based on random initialization or a pre-trained model. But such a parameter can land anywhere in some $\Theta$, so we need a uniform bound on that set. The subsequent parameters $\theta_{t} = \theta_{t-1} - \eta \nabla \hat{\cL}_n(\theta_{t-1})$ are obtained sequentially and thus has dependency on all the samples, which will be reused to estimate $\nabla \hat{\cL}_n(\theta_{t})$. Our main result shows that the deviation between the empirical gradient and the population gradient jointly over the $(T+1)$ time steps can be bounded with high probability based on the LGGW and otherwise logarithmic dependencies on the ambient dimensionality $p$ and the time horizon $T$.
\begin{restatable}{theo}{vecadagd}
\label{thm:vecada}
Let $\theta_0 \in \Theta_0$, and $\theta_t, t \in [T]$ be a sequence of parameters obtained from GD by reusing a fixed set of samples $z^{(n)} \sim \cD^n$ in each epoch. Let 
\begin{equation}
\Delta(\theta) := \left\| \frac{1}{n} \sum_{i=1}^n \nabla \ell(\theta,z_i) - \nabla \cL_D(\theta) \right\|_2~, 
\end{equation}
where the population gradient $\nabla \cL_D(\theta_t)  = \E_{z \sim \cD}[\nabla \ell(\theta_t;z)], t \in [T]$. Under Assumption~\ref{asmp:gradbnd}, with $w(\hat \Xi^0_n) $ denoting the LGGW for $\Theta_0$ as in Definition \ref{defn:gaussw}, for any $\delta \leq \frac{1}{2}$, with probability at least $(1-2\delta)$ over $z^{(n)} \sim \cD^n$,  we have 
\begin{equation}
\begin{split}
\max & \left(  \sup_{\theta_0 \in \Theta_0}  \Delta(\theta_0) ,~~ \max_{t \in [T]}  \Delta(\theta_t) \right) \leq  \frac{c_7 \max(w(\hat{\Xi}^0_n), \log p) +\sqrt{\log T + \log \frac{1}{\delta}}}{\sqrt{n}}~.
\end{split}
\label{eq:vecada}
\end{equation}
\end{restatable}
Our proof does not utilize the specific form of GD updates, and in fact goes through for any update of the form $\theta_{t+1} = f(\{\theta_{s},s \leq t\})$. Thus, Theorem \ref{thm:vecada} is an adaptive data analysis (ADA) result, but without using the traditional tools such as differential privacy~\cite{dwork_generalization_2015,JungLN0SS20}. Our proof utilizes the conditional probability of events, and expresses conditional or adaptive events in terms of events based on fresh samples. 
Proofs are in Appendix~\ref{app:sgd}. 



\begin{remark}
There are a few key takeaways from Theorem~\ref{thm:vecada}. First, note that the key term in the bound depends on LGGW $w(\hat{\Xi}^0_{n})$ only at initialization, since we considered a uniform bound over $\theta_0 \in \Theta_0$. Note that $w(\hat{\Xi}^0_{n})$ can be bounded based on properties of gradients at random initialization or with pretrained models (see Section~\ref{sec:rad_grad}). Second, the bound has a logarithmic dependence on $p$, the ambient dimension, and $T$, the number of steps. Third, the sample dependence is $\frac{1}{\sqrt{n}}$ for GD with batch size $n$. Thus, the bound says that in spite of sample reuse, the sample gradients will stay close to population gradient. \qed 
\end{remark} 
\begin{remark}
The bound in Theorem~\ref{thm:vecada} does not restrict the subsequent iterates $\theta_t$ to be restricted to the initialization set $\Theta_0$. In other words, the bound holds without a ``near initialization" type restricted which has been widely used in the deep learning theory literature. Further, in case the analysis starts from a given fixed $\theta_0$, the dependence on $w(\hat{\Xi}^0_n)$ drops out, and we are only left with dependence on $(\log p + \sqrt{\log T})$. \qed
\end{remark}

{\bf Population Convergence in Optimization.} The results in Theorems~\ref{thm:vecada} can be readily used  to get optimization
convergence of population gradients, without needing an additional uniform convergence argument~\cite{ssbd14}.

\begin{restatable}{theo}{theosgduc} \label{theo:opt_rate}
Consider a non-convex loss $\cL_{\cD}(\theta) = \E_{z\sim \cD}\left[ \ell(\theta, z) \right]$ with $\tau$-Lipschitz gradient. For GD  using a total of $z^{(n)} \sim \cD^n$ original samples and re-using the $n$ samples in each step as in Theorem~\ref{thm:vecada}, with step-size $\eta = \frac{1}{4\tau}$, with probability at least $(1-2\delta)$ for any $\delta > 0$ over $z^{(n)} \sim \cD^n$, 
 \begin{align}
 &\nr \mathbb{E}_R \left\|\nabla 
 \cL_{\cD}\left(\mathbf{\theta}_{R}\right)\right\|^{2}  \leq O\left(\frac{1}{T}\right) +  O\left(\frac{\max\left(w^{2}(\hat{\Xi}^{0}_{n}),\log^{2}p\right)+\left(\log T+\log\frac{1}{\delta}\right)}{n}\right),
 \end{align} 
where $R$ is uniformly distributed over $\{1,\ldots,T\}$ and the expectation is over the randomness of $R$.
\end{restatable}


\begin{remark}
In the bound, the first term comes from the finite-sum optimization error, which depends on the number of iterations $T$; and the second term comes from the statistical error, which depends on the sample size $n$ and the LGGW $w(\hat{\Xi}^0_{n}), \log p$, and $\log T$. \qed
\end{remark}
\begin{remark}
While an analysis of sample reuse based GD can be done using differential privacy (DP) based adaptive data analysis (ADA)~\cite{DworkFHPRR15,  JungLN0SS20} and algorithmic stability \cite{bousquet2002stability, hardt2016train}, 
the best-known efficient algorithm in such framework~\cite{bassily_differentially_2014} have gradient error scaling with $\sqrt{p}$.
Further, without any assumption, lower bounds in ADA \cite{SteinkeU16} also show that it is hard to release more than $\tilde O(n^2)$ statistical queries. Our analysis of sample reuse without using DP for the special case of estimating high-dimensional gradients leads to sharper bounds, with only logarithmic dependence on the ambient dimensionality. 
\qed 
\end{remark}

\section{Gaussian Width of Deep Learning Gradients for Neural Networks}
\label{sec:rad_grad}

%


In this section, we initiate the study of loss gradient Gaussian width (LGGW) of deep learning gradients. As a warm up, we review that bounded $L_2$ norm, e.g., $\| \xi (\theta; z) \|_2 \leq c$ for any $\theta \in \Theta$ and $z \in \cZ$, with no additional structure, leads to a large Gaussian width since for such $\hat{\Xi}_n$, 
$w({\hat{\Xi}_n}) = \E_{\g}[\sup_{\hat{\xi}_n}\langle \hat{\xi}_n, \g \rangle ] \leq c \E_{\g}[ \langle \g/\|\g\|_2, \g\rangle ] = c \E_{\g}[\|\g\|_2] = O(\sqrt{p})$. We posit that the Gaussian width of deep learning gradients is much smaller and present a set of technical results in support. We show that under standard assumptions, single sample LGGW of such deep learning models can be bounded in terms of the Gaussian width of the featurizer, i.e., output of the last-but-one layer, for both feedforward and residual networks. The results emphasize the benefits of the stability of featurizers for deep learning.

\subsection{Bounds Based on Empirical Geometry of Gradients}
\label{ssec:geometry}
\begin{figure}[t] 
\centering
 \begin{subfigure}{0.22\textwidth}
 \includegraphics[trim = 1mm 1mm 1mm 4mm, clip, width=\textwidth]{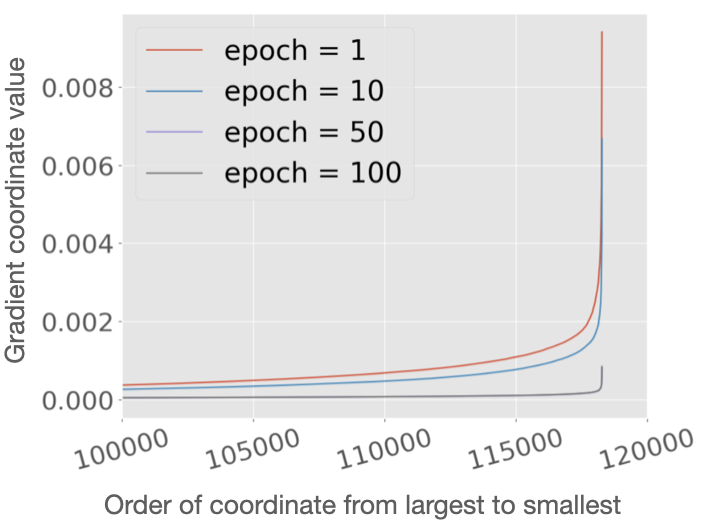}
 \caption{Sorted gradient coordinate absolute values (linear scale).}
 \end{subfigure}
 \hspace*{1mm}
 \begin{subfigure}{0.22\textwidth}
 \includegraphics[trim = 1mm 1mm 1mm 4mm, clip, width=\textwidth]{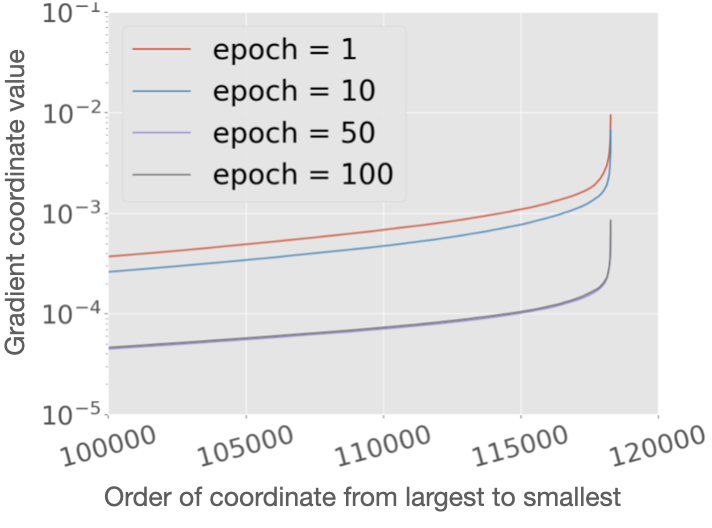}
 \caption{Sorted gradient coordinate absolute values (log scale).}
 \end{subfigure}
 \hspace*{1mm}
 \begin{subfigure}{0.22\textwidth}
 \includegraphics[trim = 1mm 1mm 1mm 4mm, clip, width=\textwidth]{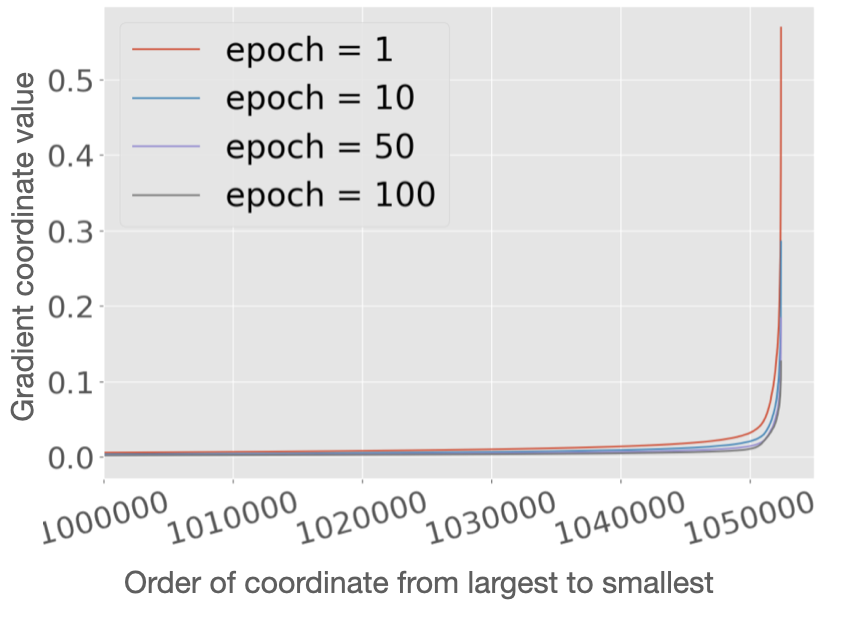}
 \caption{Sorted gradient coordinate absolute values (linear scale).}
 \end{subfigure}
 \hspace*{1mm}
 \begin{subfigure}{0.22\textwidth}
 \includegraphics[trim = 1mm 1mm 1mm 4mm, clip, width=\textwidth]{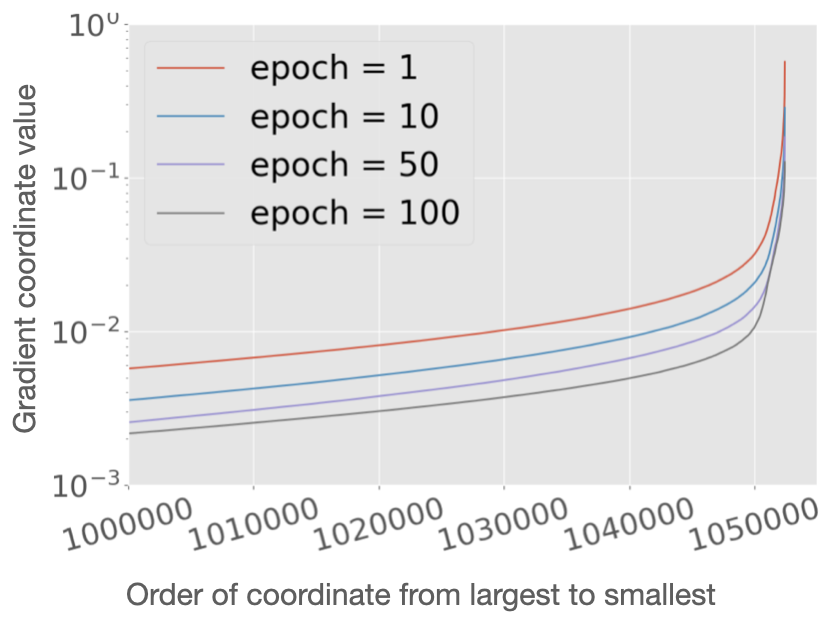}
 \caption{Sorted gradient coordinate absolute values (log scale).}
 \end{subfigure}
\caption{(a-b) Sorted gradient coordinate in linear scale (a) and log scale (b) for SGD over 100 epochs on CIFAR-10.
Model: 4-hidden-layer ReLU network with 256 nodes on each layer, with around 1,055,000 parameters. (c-d) Sorted gradient coordinate in linear scale (c) and log scale (d) for SGD over 100 epochs on MNIST. Model: 2-layer ReLU with 128 nodes on each layer, with roughly 120,000 parameters. Y-axis is the absolute value of gradient coordinates, X-axis is sorted in increasing order. }
\label{fig:cifar_eva}
\end{figure}
In the context of deep learning, recent work has illustrated that empirically the gradients are considerably structured~\cite{gur-ari_gradient_2018,papyan_full_2018,papyan_measurements_2019,ghorbani_investigation_2019,lgzcb20}: during optimization, the gradients are ``simple'' in the sense that the sorted absolute gradient components decay quickly and the components have large values only for a small subset of parameters (e.g., see Figure~\ref{fig:cifar_eva}),
and at convergence, almost all gradients are really close to zero, the so-called interpolation condition~\cite{bhmm19,negrea_defense_2019,ma_power_2017,bartlett_[1906.11300]_2019}. 
Such observations have formed the basis of more memory or communication-efficient algorithms, e.g., based on top-$k$ count-sketch, truncated SGD, and heavy hitters based analysis~\cite{rothchild2020fetchsgd,ivkin2019communication}.
%
%
We augment these empirical observations with some examples of gradient sets which have small Gaussian widths by construction or by suitable composition properties.
\begin{exam}[\bf{Ellipsoid}]
Consider $\hat{\Xi}_n = \{ \hat{\xi}_n \in \R^p \mid \sum_{j=1}^p ((\hat{\xi}_n[j])^2/a_j^2 = \tilde{O}(1), a \in \R^p\}$.
Then, the Gaussian width $w(\hat{\Xi}_n) = \tilde{O}(\| \a \|_2)$~\cite{tala14}. Note that if $\| \a \|_2 = \tilde{O}(1)$, then we have $w(\hat{\Xi}_n) = \tilde{O}(1)$. For example, if the elements of $\a$ sorted in decreasing order satisfy $a_{(j)} \leq c_3/\sqrt{j}$, then $\|\a\|_2 = \tilde{O}(1)$. \qed 
\label{exm:ellipsoid}
\end{exam}
%
%
\begin{exam}[\bf{$k$-support norm}]
The $k$-support norm 
$$\|\xi\|_k^{sp} = \inf_{\sum_i u_i = \xi}\left\{ \sum_i \| u_i \|_2 \mid \| u_i \|_0 \leq k\right\} $$ 
is based on an infimum convolution over $k$-sparse vectors~\cite{agfs12,mcdonald2014spectral,chba15}, with $k=1$ corresponding to the $L_1$ norm. Consider $\hat{\Xi}_n = \{ \hat{\xi}_n \mid \|\hat{\xi}_n\|_k^{sp} \leq c_0 \sqrt{k} \}$. Note that among many other things $\hat{\Xi}_n$ includes 
the infimum convolution of all $k$-sparse gradients $\hat{\xi}_n$ with each non-zero element bounded by $c_0$. By construction, $w(\hat{\Xi}_n) \leq c_1 k \sqrt{\log (p/k)}$~\cite{chba15}. \qed
\label{exm:ksupport}
\end{exam}

\begin{exam}[{\bf Union of sets}]
If $\hat{\Xi}_n$ is an union of $p^r$ sets, i.e., polynomial in $p$, where each set has a $\tilde{O}(1)$ Gaussian width and each element has $\tilde{O}(1)$ $L_2$ norm, then $w(\hat{\Xi}_n) = \tilde{O}(\sqrt{r})$~\cite{maurer2014inequality,chba15}. Taking convex hull of such $\hat{\Xi}_n$ can only change its Gaussian width by a constant amount, i.e., $w(\text{conv}(\hat{\Xi}_n)) \leq c_2 w(\hat{\Xi}_n)$~\cite{tala14}(Theorem 2.4.15). \qed
\label{exm:union}
\end{exam}

\begin{exam}[{\bf Minkowski sum of sets}]
If $\hat{\Xi}_n$ is Minkowski sum of a union of $p^r$ sets as in Example~\ref{exm:union} and a $\sqrt{k}$ radius $k$-support norm ball as in Example~\ref{exm:ksupport}, then $w(\hat{\Xi}_n) = \max(k,\sqrt{r})\tilde{O}(1)$. \qed 
\end{exam}

Based on how the gradients empirically look for deep networks (Figures~ \ref{fig:cifar_eva}), it is arguably fair to assume that the gradients live in $\hat{\Xi}_n$ with small Gaussian width. We prove explicit upper bounds of single-sample LGGW for FFNs and ResNets in Sections~\ref{ssec:ffn} and \ref{ssec:resnet} respectively.


\subsection{Gaussian Width Bounds for Neural Networks}
\label{ssec:gwnn}

Consider a training set $z^{(n)} = \{z_i, i \in [n]\}, z_i=(\x_i,y_i), \x_i \in \cX \subseteq \R^d, y_i \in \cY \subseteq \R$.
For some loss function $l$, the goal is to minimize:   $\hat \cL_n(\theta)  = \frac{1}{n} \sum_{i=1}^n l(  y_i, \hat{y}_i) = \frac{1}{n} \sum_{i=1}^n l (y_i,f(\theta;\x_i))$, where the prediction $\hat{y}_i:= f(\theta;\x_i)$ is from a neural network, and the parameter vector $\theta\in\R^p$. 
%
For our analysis, we will assume square loss $l(y_i,\hat{y}_i)=\frac{1}{2}(y_i-\hat{y}_i)^2$, 
though the analyses extends to more general losses under mild additional assumptions.
%
We focus on feedforward networks (FFNs) and residual networks (RestNets), and consider them as $L$ layer neural networks $f(\theta;\x)$ with parameters $\theta$, where each layer has width $m$. 
We make two standard assumptions~\cite{allen-zhu_convergence_2019,du_gradient_2019,CL-LZ-MB:20,banerjee_rsc23}.
\begin{asmp}[{\bf Activation}]
The activation function $\phi$ is 1-Lipschitz, i.e., $|\phi'| \leq 1$ and $\phi(0) = 0$.
\label{asmp:act}
\end{asmp}
\begin{asmp}[\bf{Random initialization}]
Let $\theta_0 := (\vec(W_0^{(1)})^\top,\ldots,\vec(W_0^{(L)})^\top, \v_0^\top )^\top 
$ denote the initial weights, and   $w_{0,ij}^{(l)}$ denote $i, j$-th element of  $W_0^{(l)}$ for $l\in[L]$.
At initialization, $w_{0,ij}^{(l)} \sim \cN(0,\sigma_0^{(l)}{}^2)$ for $l\in[L]$ where $\sigma_{0}^{(1)}=\frac{\sigma_{1}}{2\left(1+\sqrt{\frac{\log m_{1}}{2m_{1}}}\right)}$, $\sigma_0^{(l)} = \frac{\sigma_1}{1 + \sqrt{\frac{m_{l-1}}{m_{l}}}+\sqrt{\frac{2\log m_{l}}{m_{l}}}}$, $l\geq2$, $\sigma_1 > 0$, and 
$\norm{\v_0}_2=1$. For any input $\x \in \cX \subseteq \R^d$, $\left\|\x\right\|_{2}=1$.
\label{asmp:ginit}
\end{asmp}
Assumption \ref{asmp:act} is satisfied by commonly used activation functions such as (non-smooth) ReLU \cite{allen-zhu_convergence_2019} and (smooth) GeLU \cite{hendrycks2016gaussian}. 
And Assumption \ref{asmp:ginit} is regarding the random initialization of the weights, variants of which are typically used in most work in deep learning theory \cite{jacot_neural_2020, allen-zhu_convergence_2019,he2015delving,glorot2010understanding,sutskever2013importance,arora2019fine,du2018gradient,du_gradient_2019}.



\begin{figure*}[t]
    \centering
    \begin{subfigure}[b]{0.28\textwidth}
        \includegraphics[width=\textwidth]{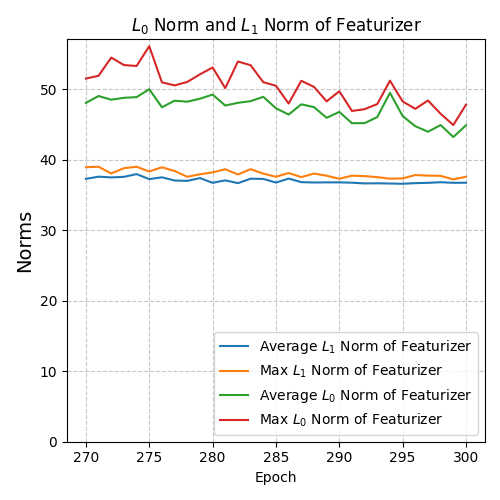}
        \caption{ResNet18 on CIFAR-10}
        \label{fig:resnet18_cifar10_feature}
    \end{subfigure}
    \hfill
    \begin{subfigure}[b]{0.28\textwidth}
        \includegraphics[width=\textwidth]{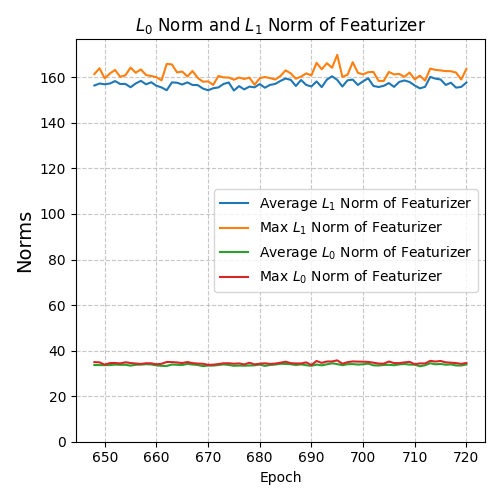}
        \caption{FFN on CIFAR-10}
        \label{fig:ffn_cifar10_feature}
    \end{subfigure}
    \hfill
    \begin{subfigure}[b]{0.28\textwidth}
        \includegraphics[width=\textwidth]{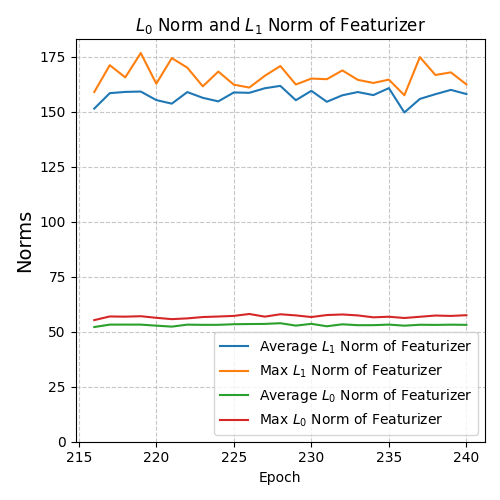}
        \caption{CNN on Fashion-MNIST}
        \label{fig:cnn_fashion_mnist_feature}
    \end{subfigure}
    \caption{$L_{0}$ norm and $L_{1}$ norm of featurizers for (a) ResNet18 on CIFAR-10, (b) FFN on CIFAR-10, and (c) CNN on Fashion-MNIST of the last 10\% epochs close to convergence. We plot both the average and maximum norms among 5 repetitions for each experiment. The figures demonstrate that for these models, as approaching convergence, the featurizers has relatively low $L_{0}$ norm and $L_{1}$ norm, which further indicates a small LGGW.}
    \label{fig:featurizer-norms}
\end{figure*}

\subsubsection{Bounds for FeedForward Networks (FFNs)}
\label{ssec:ffn}

We consider $f$ to be an FFN given by 
\begin{equation}   
f(\theta;x) = \v^\top \phi(\frac{1}{\sqrt{m_{L}}}W^{(L)} \phi(\cdots \phi( \frac{1}{\sqrt{m_{1}}}  W^{(1)} \x))))~,
 \label{eq:ffn}    
\end{equation}
where $W^{(1)} \in \R^{m_{1} \times d}, W^{(l)} \in \R^{m_{l} \times m_{l-1}}, l \in \{2,\ldots,L\}$ are the layer-wise weight matrices, $\v\in\R^{m_{L}}$ is the last layer vector, $\phi(\cdot )$ is the (pointwise) activation function, and the total set of parameters $\theta \in \R^{p}, p=\sum_{k=1}^{L}m_{k}m_{k-1}+m_{L}$ is given by
\begin{align}
\theta = (\vec(W^{(1)})^\top,\ldots,\vec(W^{(L)})^\top, \v^\top )^\top.
\label{eq:theta_def}
\end{align}
in which $m_{0}=d$.
Our analysis for both FFNs and ResNets will be for gradients over all parameters in a fixed radius {\em spectral} norm ball around the initialization $\theta_0$, which is more flexible than Frobenius norm balls typically used in the literature \cite{woodworth_kernel_2020,chizat_lazy_nodate,zou_stochastic_2018,banerjee_rsc23}:
\begin{equation}
B_{\rho, \rho_1}^{\spec}(\theta_0) := \big\{ \theta \in \R^p ~\text{as in \eqref{eq:theta_def}} ~\mid
\| \v - \v_0 \|_2 \leq \rho_1, \|W^{(\ell)} - W_0^{(\ell)} \|_2 \leq \rho, \ell \in [L]  \big\} 
\label{eq:specball} ~.
\end{equation}
Let $B_{\rho}^{\spec}(\theta_0) \in \R^{p-m_{L}}$ be the corresponding set of weight matrices $W = \{W^{(l)}, \ell \in [L]\}$. The main implications of our results and analyses go through as long as the layerwise spectral radius $\rho = o(\sqrt{m_{l}})$ for all $l\in[L]$, and last layer radius $\rho_1 = O(1)$, which are arguably practical. 

Let 
 \begin{align}
 h^{(L)}(W,\x) & := \phi\left(\frac{1}{\sqrt{m_{L}}}W^{(L)} \phi(\cdots \phi( \frac{1}{\sqrt{m_{1}}}  W^{(1)} \x))) \right) ~\in \R^{m_{L}}
 \end{align}
be the ``featurizer,'' i.e., output of the last but one layer, so that $f(\theta;\x) = \v^\top h^{(L)}(W,\x)$ with $\theta = (W,\v)$. Consider the set of all featurizer values for a single sample:
\begin{align}\label{eq:feature_set}
A^{(L)}(\x) = \left\{ h^{(L)}(W,\x) \mid W \in B_{\rho}^{\spec}(\theta_0) \right\}
\end{align}  

We now establish an upper bound on single-sample LGGW 
of FFN gradient set $\Xi^{\ffn}$ corresponding to parameters in $B_{\rho, \rho_1}^{\spec}(\theta_0)$. In particular, we show that the single-sample LGGW 
can effectively be bounded by the Gaussian width of the featurizer set.
\begin{restatable}[\bf{LGGW: FFNs}]{theo}{theofnngradwidth}
\label{cor:gradient-bounds}
Under Assumptions~\ref{asmp:act} and \ref{asmp:ginit}, with $\beta_{l} := \sigma_1 + \frac{\rho}{\sqrt{m_{l}}}$, $l\in[L]$, with probability at least $\left(1-\sum_{l=1}^{L}\frac{2}{m_{l}}\right)$ over the randomness of the initialization, we have 
\begin{align}
w(\Xi^{\ffn}) \leq c_{1}w(A^{(L)})+c_{2}(1+\rho_{1})\sqrt{m_{L}}\left(\prod_{l=1}^{L}\beta_{l} \right) \sum_{l=1}^{L}\frac{1}{\beta_{l}\sqrt{m_{l}}}~.
\end{align}
\end{restatable}

\begin{remark}
For the simplest case, with $m_{l}=m$ for all $l\in[L]$, we have that
\begin{align*}w(\Xi^{\ffn}) \leq c_1 w(A^{(L)} ) + c_2 L(1+\rho_{1}) \beta^{L-1},
\end{align*}
where $\beta = \sigma_{1}+\frac{\rho}{\sqrt{m}}$. Note that choosing $\sigma_1 < 1 - \frac{\rho}{\sqrt{m}}$, i.e., mildly small initialization variance, satisfies $\beta < 1$. When $\beta < 1$, we have $w(\Xi^{\ffn}) = O(1)$ as long as $L = \Omega ( \frac{\log (\sqrt{m} + L(1+\rho_{1}))}{\log \frac{1}{\beta}} )$,
which holds for moderate depth $L$. To see this, note that we have $w(\Xi^{\ffn}) = O(w(A^{(L)}) + L (1+\rho_{1})\beta^{L-1})$.  Further, we have whp $\| \alpha^{(L)} \|_2 \leq \beta^L$ (Lemma~\ref{lemma:layer} in Appendix~\ref{app:gauss_width}), 
so that $w(A^{(L)}) \leq c \beta^L \sqrt{m}$. Then, for $\beta < 1$, the bound $\beta^L \sqrt{m} + \beta^{L-1} L(1+\rho_{1}) = O(1)$ if $L = \Omega ( \frac{\log (\sqrt{m} + L(1+\rho_{1}))}{\log \frac{1}{\beta}} )$. Interestingly, sufficient depth $L$ helps control the Gaussian width.
\qed 
\end{remark}
\begin{remark}
In practice, the dimensions of the hidden layers $m_{l}, l\in[L-1]$ are always much larger than the dimension of the output layer $m_{L}$, i.e., $m_l \gg m_L$, and our bound is even better under such situations. To see this, first note that $1- \sum_{l=1}^{L}\frac{2}{m_{l}}\gg 1 - \frac{2L}{m_{L}}$. 
Second, for the second term in the bound which depends on parts other than the featurizer, $  \sqrt{m_{L}}\prod_{l=1}^{L}\beta_{l}\sum_{l=1}^{L}\frac{1}{\beta_{l}\sqrt{m_{l}}}\ll L \beta_{L}^{L-1}$, since $\beta_L > \beta_l$.
Therefore, the probability is higher and the bound is smaller when $m_l \gg m_L$, as in real neural networks.
\qed
\end{remark}
\begin{remark}
The result implies that under suitable conditions, the Gaussian width of gradients over all $p = \sum_{k=1}^{d}m_{k}m_{k-1}+m_{L}$ parameters can be reduced to the Gaussian width of the featurizer, which is $m_{L}$ dimensional. If the featurizer does feature selection, e.g., as in variants of invariant risk minimization (IRM)~\cite{arjovsky_invariant_2020,zhou_sparse_2022} or sharpness aware minimization (SAM)~\cite{foret2020sharpness,andriushchenko_sharpness-aware_2023}, then the Gaussian width of the featurizer will be small. On the other hand, if the featurizer has many spurious and/or random features~\cite{rosenfeld_risks_2021,zhou_sparse_2022}, then the Gaussian width will be large. \qed 
\end{remark}


\subsubsection{Bounds for Residual Networks (ResNets)}
\label{ssec:resnet}

We consider $f$ to be a ResNet  given by 
 \begin{equation}
 \begin{split}
  \alpha^{(l)}(\x)  & =  \alpha^{(l-1)}(\x) + \phi \left( \frac{1}{L \sqrt{m_{l}}} W^{(l)} \alpha^{(l-1)}(\x) \right)~,\\
 f(\theta;\x) & = \alpha^{(L+1)}(\x)  = \v^\top \alpha^{(L)}(\x)~,
 \end{split}
 \label{eq:resnet}    
 \end{equation}
where $\alpha^{(0)}(\x) = \x$, and $W^{(l)}, l \in [L], \v, \phi$ are as in FFNs. Let $h^{(L)}(W,\x)  := \alpha^{(L)}(\x)\in\mathbb{R}^{m_{L}}$ be the ``featurizer,'' i.e., output of the last layer, so that $f(\theta;\x) = \v^\top h^{(L)}(W,\x)$ with $\theta = (W,\v)$.
Note that the scaling of the residual layers is assumed to have a smaller scaling factor $\frac{1}{L}$ following standard theoretical analysis of ResNets~\cite{du_gradient_2019} and also aligns with practice where smaller initialization variance yields state-of-the-art performance~\cite{zhang_understanding_2017}.
With the featurizer and associated sets defined as in \eqref{eq:feature_set} for FFNs, we now establish an upper bound on the 
single-sample LGGW of ResNet gradients in terms of that of the featurizer.
\begin{restatable}[\bf{LGGW: ResNets}]{theo}{theorngradwidth}
\label{cor:gradient-bounds-resnet}
Under Assumptions~\ref{asmp:act} and \ref{asmp:ginit}, with $\beta_{l} := \sigma_1 + \frac{\rho}{\sqrt{m_{l}}}$, $l\in[L]$, with probability at least $\left(1-\sum_{l=1}^{L}\frac{2}{m_{l}}\right)$ over the randomness of the initialization,  we have 
\begin{align}
w(\Xi^{\resnet}) \leq c_{1}w(A^{(L)})+c_{2}\frac{1+\rho_{1}}{L}\sqrt{m_{L}}\prod_{l=1}^{L}\left(1+\frac{\beta_{l}}{L}\right)\sum_{l=1}^{L}\frac{1}{\left(1+\frac{\beta_{l}}{L}\right)\sqrt{m_{l}}}.
\end{align}
\end{restatable}
\begin{remark}
For the simplest case, with $m_{l}=m$ for all $l\in[L]$, we have that $w(\Xi^{\resnet}) \leq c_1 w(A^{(L)}) + c_{2}\left(1+\rho_{1}\right)\left(1+\frac{\beta}{L}\right)^{L-1}
    \leq c_1 w(A^{(L)}) + c_2 (1+\rho_{1})e^{\beta}$.
in which $\beta = \sigma_{1}+\frac{\rho}{\sqrt{m}}$, and the second inequality holds since $\left(1+\frac{\beta}{L}\right)^{L-1}\leq e^{\beta}$. As for FFNs, choosing $\sigma_1 < 1 - \frac{\rho}{\sqrt{m}}$, i.e., mildly small initialization variance, satisfies $\beta < 1$. When $\beta < 1$, we have $w(\Xi^{\resnet}) = O( w(A^{(L)}) )$ since $(1+\rho_{1})e^\beta = O(1)$. Thus, for ResNets, the LGGW on all parameters reduces to the Gaussian width of the featurizer.  \qed 
\end{remark}
\begin{remark}
As for FFNs, our ResNet bound is also much sharper when the dimensions in the hidden layers are much larger than the output dimension, which is the case in commonly used ResNet models. 
\qed
\end{remark}
\begin{remark}
To get a sense of the Gaussian width of the featurizer, which will be small if the featurizer is sparse~\cite{vershynin_high-dimensional_2018}, in Figure \ref{fig:featurizer-norms}, we plot the average and max (over 5 runs) of the $L_0$ and $L_{1}$ norm of the featurizer for the last $10\%$ epochs close to convergence for three standard deep learning models: ResNet18 on CIFAR-10 (512 dimension featurizer), FFN on CIFAR-10 (256 dimension featurizer), and CNN on Fashion-MNIST (128 dimension featurizer). 
The small $L_0$ and $L_1$ norm of the featurizer, indicating both feature selection and being in small $L_1$ ball, imply small Gaussian width of the featurizer which in turn implies small LGGW based on Theorems~\ref{cor:gradient-bounds} and \ref{cor:gradient-bounds-resnet}.
\qed  
\end{remark}

\section{Conclusions}
\label{sec:conc}
In this work, we have introduced a new approach to analyzing the generalization and optimization of learning models based on the loss gradient Gaussian width (LGGW). For loss satisfying the gradient domination condition, which includes the popular PL condition as a special case, we establish generalization bounds in terms of LGGW. In the context of optimization, we show that gradient descent with sample reuse does not go haywire and the gradient estimates stay close to the population counterparts as long as the LGGW at initialization is small. We also consider the LGGW of deep learning models, especially feedforward networks and residual networks. In these settings, we show that the LGGW with single sample can be bounded by the Gaussian width of the featurizer and additional terms which are small under mild conditions. Overall, our work demonstrates the benefits of small loss gradient Gaussian width (LGGW) 
in the context of learning, and will hopefully serve as motivation for further study of this geometric perspective especially for modern learning models.


\paragraph{Acknowledgements.} The work was supported in part by the National Science Foundation (NSF) through
awards IIS 21-31335, OAC 21-30835, DBI 20-21898, as well as a C3.ai research award.


\bibliographystyle{plain}

\newpage
\appendix

\section{Generalization Bounds: Proofs for Section~\ref{sec:gen}}
\label{app:gen}

In this section, we first provide a proof sketch of the main results. Then we provide detail proofs for results in Section~\ref{sec:gen}.

\subsection{Generalization bound in terms of NERC}

We start with a variation of Proposition 1 of \cite{FSS}, which will be convenient for our purposes.
\begin{prop}
\label{prop:genprop1}
Under Assumptions~\ref{asmp:gd_cond} and \ref{asmp:gradbnd}, with $\theta^{\star} \in \arg \min _{\theta \in \Theta} L_{\mathcal{D}}(w)$ denoting any population loss minimizer and $\hat{R}_n(\hat{\Xi}^{(n)})$ as in Definition \ref{defn:nerc}, for any $\delta > 0$, 
with probability at least $1-\delta$ over the draw of the samples $z^{(n)} \sim \cD^n$, for any $\hat{\theta} \in \Theta$ we have
\begin{align}
 \cL_{\cD}(\hat{\theta}) - \cL(\theta^*)   \leq 2c_0 \left( \| \nabla \hat{\cL}_n(\hat{\theta}) \|^{\alpha} + 2\left( \E \sup_{\theta \in \Theta} \| \nabla \hat{\cL}_n(\theta) - \nabla \cL_D(\theta) \| \right)^{\alpha} + c_2 \left( \frac{\log \frac{1}{\delta}}{n} \right)^{\frac{\alpha}{2}} \right) ~.
\end{align}
\end{prop}
\proof Since $\cL_{\cD}$ satisfies the $(\alpha,c_0)$-GD condition, since $\alpha \in [1,2]$, we have 
\begin{align*}
 \cL_{\cD}(\hat{\theta}) - \cL(\theta^*) & \leq c_0 \| \nabla \cL_{\cD}(\hat{\theta}) \|^{\alpha}
  \leq 2c_0 \left( \| \nabla \hat{\cL}_{n}(\hat{\theta}) \|^{\alpha} + \| \nabla \hat{\cL}_{n}(\hat{\theta}) - \nabla \cL_{\cD}(\hat{\theta}) \|^{\alpha} \right) \\
  & \leq 2c_0 \left( \| \nabla \hat{\cL}_{n}(\hat{\theta}) \|^{\alpha} + \left( \sup_{\theta \in \Theta} \| \nabla \hat{\cL}_{n}(\theta) - \nabla \cL_{\cD}(\theta) \| \right)^{\alpha} \right)
\end{align*}
Now, by a direct application of McDiarmid's inequality, since $z^{(n)} \sim \cD^n$, with probability at least $(1-\delta)$ over the draw of $z^{(n)}$, we have
\begin{align*}
 \sup_{\theta \in \Theta}~\left\| \frac{1}{n} \sum_{i=1}^n \nabla \ell(\theta;z_i) - \E_z[\nabla \ell(\theta;z)] \right\|
 \leq \E \sup_{\theta \in \Theta} \left\| \frac{1}{n} \sum_{i=1}^n \nabla \ell(\theta;z_i) - \E_z[\nabla \ell(\theta;z)] \right\| + c \sup_{\substack{\theta \in \Theta \\z \in \cZ}} \| \nabla \ell(\theta;z) \| \sqrt{ \frac{\log \frac{1}{\delta}}{n}} ~.
\end{align*}
Noting that $\|\nabla \ell(\theta;z)\|_2 = O(1)$ for $\theta \in \Theta, z \in \cZ$ completes the proof. \qed 

Next, we effectively restate Proposition 2 of \cite{FSS}, in our notation. 
\begin{prop} 
\label{prop:rc_uc}
Let $\hat{\Xi}^{(n)}$ be the set of all $n$-tuple gradients as in \eqref{eq:hatxin} and $\hat{R}_n(\hat{\Xi}^{(n)})$ be the NERC (normed empirical Rademacher complexity) as in Definition~\ref{defn:nerc}. 
For any $\delta >0$, with probability at least $1-\delta$ over the draw of $z^{(n)} \sim \cD^n$, we have
\begin{equation}
\mathbb{E} \sup _{\theta \in \Theta}\left\|\nabla \widehat{L}_{n}(\theta)-\nabla L_{\mathcal{D}}(\theta)\right\| \leq 4 \cdot \hat{R}_n(\hat{\Xi}^{(n)}) +c_1 \left(\frac{\log \left(\frac{1}{\delta}\right)}{n}\right) ~.
\end{equation}
\end{prop}
\proof The proof follows from that of Proposition 2 of \cite{FSS}. \qed 

Combining Propositions~\ref{prop:genprop1} and \ref{prop:rc_uc}, we now get the following result.
\genbnd*

\subsection{Bounding NERC with Gaussian width}

\vechf*

We establish the result by an application of the entropy method, and the specific application is just the bounded difference inequality~\cite{boucheron_concentration_2013}. Recall that a function $f : \cX^n \mapsto \R$ is said to have the bounded difference property is for some non-negative constants $c_1,\ldots,c_n$, we have for $1\leq i \leq n$
\begin{equation}
\sup_{\substack{x_1,\ldots,x_n\\x_i'\in \cX}}~|f(x_1,\ldots,x_n) - f(x_1,\ldots,x_{i-1},x_i',x_{i+1},\ldots,x_n)| \leq c_i~.
\end{equation}
If $f$ satisfies the bounded difference property, then an application of the Efron-Stein inequality~\cite{boucheron_concentration_2013}[Theorem 3.1] implies that $Z = f(X_1,\ldots,X_n)$ satisfies the variance $\text{var}(Z) \leq \frac{1}{4}\sum_{i=1}^n c_i^2$~\cite{boucheron_concentration_2013}[Corollary 3.2]. The bounded difference inequality is an application of the entropy method~\cite{boucheron_concentration_2013}[Theorem 6.2] which shows that such $Z$ satisfies a sub-Gaussian tail inequality where the role of the sub-Gaussian norm~\cite{vers12,vershynin_high-dimensional_2018} is played by the variance $v=\frac{1}{4} \sum_i c_i^2$, so that
\begin{equation}
    \P(|Z - \E[Z]| > u ) \leq 2 \exp(-u^2/2v)~.
\label{eq:bndiff}
\end{equation}
We now focus on the specific result we need and prove it using the bounded difference inequality.

\proof Since $\v_i$ are fixed, let 
\begin{align*}
f(\eps_i,\ldots,\eps_n) := \left\| \frac{1}{\sqrt{n}} \sum_{i=1}^n \eps_i \v_i \right\|_2 ~.
\end{align*}
By triangle inequality, for any $j \in \{1,\ldots,n\}$, we have
\begin{align*}
\sup_{\substack{\eps_1,\ldots,\eps_n\\\eps_j'\in \{-1,+1\}}} &~|f(\eps_1,\ldots,\eps_n) - f(\eps_1,\ldots,\eps_{j-1},\eps_j',\eps_{j+1},\ldots,\eps_n)| \\
& \leq \sup_{\substack{\eps_1,\ldots,\eps_n\\\eps_j'\in \{-1,+1\}}} ~\left| \left\| \frac{1}{\sqrt{n}} \sum_{i=1}^n \eps_i \v_i \right\|_2  - \left\| \frac{1}{\sqrt{n}} \left( \sum_{i=1}^n \eps_i \v_i  - \eps_j \v_j + \eps_j' \v_j \right) \right\|_2 \right| \\
& \leq \sup_{\eps_j,\eps_j' \in \{-1,+1\}} \frac{1}{\sqrt{n}}  \left\| \eps_j \v_j - \eps_j' \v_j \right\|_2\\
& \leq \frac{2}{\sqrt{n}} \| \v_j \|_2 ~.
\end{align*}
As a result, the varinace bound $v = \frac{1}{4} \sum_{i=1}^n \frac{2^2}{n} \| \v_i \|_2^2 \leq 1$, since $\sum_{i=1}^n \| \v_i \|^2_2 \leq n$.

Further, we have $\E[f(\eps_1,\ldots,\eps_n)] \leq 1$. To see this, first note that
\begin{align*}
\E[f^2(\eps_1,\ldots,\eps_n)] = \E \left[ \left\| \frac{1}{\sqrt{n}} \sum_{i=1}^n \eps_i \v_i \right\|_2^2 \right] 
 = \frac{1}{n} \E \left[ \sum_{i=1}^n \| \v_i \|_2^2 + \sum_{\substack{i,j=1 \\ i \neq j}}^n \langle \eps_i \v_i, \eps_j \v_j \rangle \right] \overset{(a)}{\leq} 1
\end{align*}
where $(a)$ follows since $\sum_{i=1}^n \| \v_i \|^2_2 \leq n$ and by independence, for $i \neq j$, $\E[ \langle \eps_i \v_i, \eps_j \v_j \rangle] = \langle \E[\eps_i \v_i], \E[\eps_j \v_j] \rangle = 0$ since $\E[\eps_i] = 0$. Then, by Jensen's inequality, we have
\begin{align*}
\E[f((\eps_1,\ldots,\eps_n)] & = \E \left[ \left\| \frac{1}{\sqrt{n}} \sum_{i=1}^n \eps_i \v_i \right\|_2 \right] 
 \leq \sqrt{ \E \left[ \left\| \frac{1}{\sqrt{n}} \sum_{i=1}^n \eps_i \v_i \right\|_2^2 \right]  }  \leq 1~. 
\end{align*}
Then, an application of the bounded difference inequality~\eqref{eq:bndiff}~\cite{boucheron_concentration_2013}[Theorem 6.2] completes the proof. \qed

Next we focus on the proof of Lemma~\ref{thm:shiftgc}. The proof follows the standard generic chaining analysis~\cite{tala14} with suitable adjustments to handle the constant shift $\mu$.

\shiftgc*

\proof We consider an optimal admissible sequence $\{\Gamma^{(n)}_r\}$, i.e., sequence of subsets $\Gamma^{(n)}_r$ of ~$\hat{\Xi}^{(n)}$ with $|\Gamma^{(n)}_r| \leq N_{r}$,
where  $N_{0}=1, N_{r}=2^{2^{r}} \text { if } r \geq 1$, and 
\begin{align}
\gamma_2(\hat{\Xi}^{(n)},\bar{d}^{(n)}) = \sup_{\hat{\xi}^{(n)} \in \hat{\Xi}^{(n)}} \sum_{r \geq 0} 2^{r/2} \bar{d}^{(n)}(\hat{\Xi}^{(n)},\Gamma^{(n)}_r)~.
\end{align}
For any $\hat{\xi}^{(n)} \in \hat{\Xi}^{(n)}$, let $\pi_r(\hat{\xi}^{(n)}) \in \Gamma^{(n)}_r$ be the "projection" of $\hat{\xi}^{(n)}$ to $\Gamma^{(n)}_r$, i.e., 
\begin{align}
\pi_r(\hat{\xi}^{(n)}) = \underset{\hat{\xi}^{(n)}_r  \in\Gamma^{(n)}_r}{\argmin} ~\bar{d}^{(n)}(\hat{\xi}^{(n)},\hat{\xi}^{(n)}_r)~.
\end{align}
Further, let $\hat{\xi}^{(n)}_0 \in \hat{\Xi}^{(n)}$ be such that $X_{\hat{\xi}^{(n)}_0} = 0$. For any $\hat{\xi}^{(n)} \in \hat{\Xi}^{(n)}$, we can decompose $X_{\hat{\xi}^{(n)}}$ as
\begin{align} \label{eq:chain}
 X_{\hat{\xi}^{(n)}} = \sum_{r \geq 1}\left(X_{\pi_{r}(\hat{\xi}^{(n)})}-X_{\pi_{r-1}(\hat{\xi}^{(n)})}\right) + X_{\hat{\xi}^{(n)}_{0}}~,
\end{align}
which holds provided we have chosen the sequence of sets $\{\Gamma^{(n)}_r\}$ such that $\pi_{R}(\hat{\xi}^{(n)})=\hat{\xi}^{(n)}$ for $R$ large enough.  Now, based on the shifted increment condition \eqref{eq:ic}, for any $u>0$, we have
\begin{align}
\P( \left|X_{\pi_{r}(\hat{\xi}^{(n)})}- X_{\pi_{r-1}(\hat{\xi}^{(n)})}\right| >  c_0 (\mu + u 2^{r/2}) \bar{d}^{(n)}(\pi_{r}(\hat{\xi}^{(n)}),\pi_{r-1}(\hat{\xi}^{(n)}) ) \leq c_1 \exp(-u^22^{r}/2) ~.
\end{align}
Over all $\hat{\xi}^{(n)} \in \hat{\Xi}^{(n)}$, the number of possible pairs $\left(\pi_{r}(\hat{\xi}^{(n)}), \pi_{r-1}(\hat{\xi}^{(n)})\right)$ is 
\begin{equation*}
|\Gamma^{(n)}_r| \cdot |\Gamma^{(n)}_{r-1}| \leq N_{r} N_{r-1} \leq N_{r+1}=2^{2^{r+1}}~.
\end{equation*}
Applying union bound over all the possible pairs of $\left(\pi_{r}(\hat{\xi}^{(n)}), \pi_{r-1}(\hat{\xi}^{(n)})\right)$, we have
\begin{align} 
\forall r \geq 1, \forall \hat{\xi}^{(n)} \in \hat{\Xi}^{(n)}, \qquad \left|X_{\pi_{r}(\hat{\xi}^{(n)})}-X_{\pi_{r-1}(\hat{\xi}^{(n)})}\right| \leq c_0 (\mu + u 2^{r/2}) \bar{d}^{(n)} (\pi_{r}(\hat{\xi}^{(n)}),\pi_{r-1}(\hat{\xi}^{(n)}))
\label{eq:bnd2}
\end{align}
with probability at least
\begin{align}
1 - \sum_{r\geq 1} 2^{2^{r+1}} \cdot c_1 \exp(-u^22^{r}/2)  \geq 1 - c_3 \exp(-u^2/2)~,
\label{un_ic}
\end{align}
where $c_3$ is a positive constant. 
From \eqref{eq:chain}, \eqref{eq:bnd2}, and \eqref{un_ic}, using the fact that $X_{\hat{\xi}^{(n)}_0}=0$, we have 
\begin{align*}
  \P & \left(  \sup_{\hat{\xi}^{(n)} \in \hat{\Xi}^{(n)}}~ \left| X_{\hat{\xi}^{(n)}} \right| > \sup_{\hat{\xi}^{(n)} \in \hat{\Xi}^{(n)}} \sum_{r \geq 1} c_0 (\mu + u 2^{r/2})   \bar{d}^{(n)}(\pi_{r}(\hat{\xi}^{(n)}), \pi_{r-1}(\hat{\xi}^{(n)}))  \right) \leq c_3 \exp \left(-\frac{u^{2} }{2}\right)~.
\end{align*}
By triangle inequality, we have 
\begin{align*}
  \bar{d}^{(n)}(\pi_{r}(\hat{\xi}^{(n)}),\pi_{r-1}(\hat{\xi}^{(n)})) \leq   \bar{d}^{(n)}(\hat{\xi}^{(n)},\pi_{r}(\hat{\xi}^{(n)})) + \bar{d}^{(n)}(\hat{\xi}^{(n)},\pi_{r-1}(\hat{\xi}^{(n)}))~,
\end{align*}
so that
\begin{align*}
\sup_{\hat{\xi}^{(n)} \in \hat{\Xi}^{(n)}} \sum_{r \geq 1} (\mu + u 2^{r/2}) \bar{d}^{(n)}(\pi_{r}(\hat{\xi}^{(n)}),\pi_{r-1}(\hat{\xi}^{(n)})) 
& \leq \sup_{\hat{\xi}^{(n)} \in \hat{\Xi}^{(n)}} 2 \sum_{r \geq 0} (\mu + u 2^{r/2}) \bar{d}^{(n)}(\hat{\xi}^{(n)},\pi_{r}(\hat{\xi}^{(n)})) \\
& \leq 2  (\mu + u) \sup_{\hat{\xi}^{(n)} \in \hat{\Xi}^{(n)}} \sum_{r \geq 0} 2^{r/2} \bar{d}^{(n)}(\hat{\xi}^{(n)},\pi_{r}(\hat{\xi}^{(n)})) \\
& \leq 2 (\mu + u) \gamma_2(\hat{\Xi}^{(n)},\bar{d}^{(n)})~.
\end{align*}
As a result, we have 
\begin{align}\label{eq:ch_bound0}
\mathrm{P}\left(\sup _{\hat{\xi}^{(n)} \in \hat{\Xi}^{(n)}}  \left| X_{\hat{\xi}^{(n)}} \right| > 2 c_0 (\mu + u) \gamma_2(\hat{\Xi}^{(n)},\bar{d}^{(n)})\right)  \leq
c_3 \exp \left(-\frac{u^{2}}{2}\right).
\end{align}

That completes the proof. \qed


Using Lemma~\ref{thm:vechf} and \ref{thm:shiftgc}, we can now prove Theorem~\ref{theo:rcgc}.

\rcgc*

\proof For any $\hat{\xi}^{(n)}_1, \hat{\xi}^{(n)}_2 \in \hat{\Xi}^{(n)}$, for any $u > 0$, by triangle inequality,
\begin{align*}
\hspace*{-5mm}  \P_{\eps^{(n)}} & \left( \left\| \frac{1}{\sqrt{n}} \sum_{i=1}^n \eps_i \xi_{i,1} \right\|_2 - \left\| \frac{1}{\sqrt{n}} \sum_{i=1}^n  \eps_i \xi_{i,2} \right\|_2 
 \geq  u \bar{d}^{(n)}(\hat{\xi}_{1}^{(n)},\hat{\xi}_{2}^{(n)}) \right) 
 \leq \P_{\eps^{(n)}} \left( \left\| \frac{1}{\sqrt{n}} \sum_{i=1}^n \eps_i (\xi_{i,1} - \xi_{i,2}) \right\|_2   \geq u \bar{d}^{(n)}(\hat{\xi}_{1}^{(n)},\hat{\xi}_{2}^{(n)}) \right) ~.
\end{align*}
Similarly,
\begin{align*}
\hspace*{-5mm} \P_{\eps^{(n)}} & \left( \left\| \frac{1}{\sqrt{n}} \sum_{i=1}^n \eps_i \xi_{i,2} \right\|_2 - \left\| \frac{1}{\sqrt{n}} \sum_{i=1}^n  \eps_i \xi_{i,1} \right\|_2 
 \geq  u \bar{d}^{(n)}(\hat{\xi}_{1}^{(n)},\hat{\xi}_{2}^{(n)}) \right) 
 \leq \P_{\eps^{(n)}} \left( \left\| \frac{1}{\sqrt{n}} \sum_{i=1}^n \eps_i (\xi_{i,1}- \xi_{i,2}) \right\|_2   \geq u \bar{d}^{(n)}(\hat{\xi}_{1}^{(n)},\hat{\xi}_{2}^{(n)}) \right)  ~.
\end{align*}
Let $\v_i = \frac{\xi_{i,1} - \xi_{i,2}}{\bar{d}^{(n)}(\hat{\xi}_{1}^{(n)},\hat{\xi}_{2}^{(n)})}$, so that $\sum_{i=1}^n \|\v_i \|^2_2 = n$. Then, from Lemma~\ref{thm:vechf}, for some $\mu \leq 1$, for any $u > 0$ we have
\begin{align*}
    \P_{\eps^{(n)}} \left( \left| \left\| \frac{1}{\sqrt{n}} \sum_{i=1}^n \eps_i \v_i \right\|_2 - \mu \right| \geq u \right) 
    & \leq 2\exp \left(- \frac{u^2}{2}\right) ~.
\end{align*}
so that
\begin{align*}
\P_{\eps^{(n)}} \left( \left\| \frac{1}{\sqrt{n}} \sum_{i=1}^n \eps_i (\xi_{i,1} - \xi_{i,2})  \right\|_2 \geq ( \mu +u) \bar{d}^{(n)}(\hat{\xi}_{1}^{(n)},\hat{\xi}_{2}^{(n)}) \right)
    & \leq 2\exp \left(- \frac{u^2}{2}\right) 
\end{align*}
Then,
\begin{align*}
\P_{\eps^{(n)}} & \left( \left| \left\| \frac{1}{\sqrt{n}} \sum_{i=1}^n \eps_i \xi_{i,1} \right\|_2 - \left\| \frac{1}{\sqrt{n}} \sum_{i=1}^n  \eps_i \xi_{i,2} \right\|_2 \right| \geq  (\mu + u) \bar{d}^{(n)}(\hat{\xi}_{1}^{(n)},\hat{\xi}_{2}^{(n)}) \right) \\ 
& \leq  \P_{\eps^{(n)}} \left( \left\| \frac{1}{\sqrt{n}} \sum_{i=1}^n \eps_i \xi_{i,1} \right\|_2 - \left\| \frac{1}{\sqrt{n}} \sum_{i=1}^n  \eps_i \xi_{i,2} \right\|_2 
 \geq  (\mu + u) \bar{d}^{(n)}(\hat{\xi}_{1}^{(n)},\hat{\xi}_{2}^{(n)}) \right) \\
 & \qquad \qquad +  \P_{\eps^{(n)}} \left( \left\| \frac{1}{\sqrt{n}} \sum_{i=1}^n \eps_i \xi_{i,1} \right\|_2 - \left\| \frac{1}{\sqrt{n}} \sum_{i=1}^n  \eps_i \xi_{i,2} \right\|_2 
 \geq  (\mu + u) \bar{d}^{(n)}(\hat{\xi}_{1}^{(n)},\hat{\xi}_{2}^{(n)}) \right) \\
& \leq 2 \P_{\eps^{(n)}} \left( \left\| \frac{1}{\sqrt{n}} \sum_{i=1}^n \eps_i (\xi_{i,1} - \xi_{i,2})  \right\|_2 \geq (\mu+u)   \bar{d}^{(n)}(\hat{\xi}_{1}^{(n)},\hat{\xi}_{2}^{(n)}) \right) \\
& \leq 4 \exp(-u^2/2)~.
\end{align*}

 

Then, applying Lemma~\ref{thm:shiftgc} with constants, for any $u > 0$ we have
\begin{align}
    \P_{\eps^{(n)}}\left[ \sup_{\hat{\xi}^{(n)} \in \hat{\Xi}^{(n)}} \frac{1}{\sqrt{n}} \left\| \sum_{i=1}^n \eps_i \xi_i \right\|_2 \geq c_1 (1 + u)\gamma_2(\hat{\Xi}^{(n)},\bar{d}^{(n)}) \right] \leq c_0 \exp(-u^2/2)~.
\end{align}
Let 
\begin{align*}
Y = \frac{\underset{{\hat{\xi}^{(n)} \in \hat{\Xi}^{(n)}}}{\sup}~ \frac{1}{\sqrt{n}}\| \sum_{i=1}^n \eps_i \xi_i\|_2}{c_1 \gamma_2(\hat{\Xi}^{(n)},\bar{d}^{(n)})}~. 
\end{align*}
For any constant $u_0 > 0$, i.e., $u_0 = O(1)$, we have
\begin{align*}
     \mathbb{E} [Y] & = \int_{0}^{\infty} \P(Y \geq u) \mathrm{d} u \\
     & = \int_{0}^{1+u_0} \P (Y \geq u) \mathrm{d} u + \int_{u_0 + 2}^{\infty} \P(Y \geq u) \mathrm{d} u\\
     & \leq 1+ u_0  + \int_{u_0 }^{\infty} \P(Y \geq 2 + u) \mathrm{d} u \\
    & = 1 + u_0 + \frac{c_0}{u_0} \exp\left(-\frac{u_0^2}{2}\right) \\
    & \leq 1 + c_3,
\end{align*}
where $c_3$ is a constant. Hence,
\begin{align*}
\E_{\eps^{(n)}} \left[ \sup _{\hat{\xi}^{(n)} \in \hat{\Xi}^{(n)}} \frac{1}{\sqrt{n}} \left\| \sum_{i=1}^n \eps_i \xi_i  \right\|_2 \right] & \leq c_1 (1 + c_3) \gamma_2(\hat{\Xi}^{(n)},\bar{d}^{(n)})~ \\
\Rightarrow \qquad \frac{1}{n} \E_{\eps^{(n)}} \left[ \sup _{\hat{\xi}^{(n)} \in \hat{\Xi}^{(n)}} \left\| \sum_{i=1}^n \eps_i \xi_i  \right\|_2 \right] & \leq c_2 \frac{\gamma_2(\hat{\Xi}^{(n)},\bar{d}^{(n)})}{\sqrt{n}}~,
\end{align*} 
for a suitable constant $c_2$. 

Now that we have established a version of both parts of Theorem~\ref{theo:rcgc} in terms of $\gamma_2(\hat{\Xi}^{(n)},\bar{d}^{(n)})$, to complete the proof by showing that $\gamma_2(\hat{\Xi}^{(n)},\bar{d}^{(n)}) \leq c_4 w(\hat{\Xi}_n)$ for some constant $c_4$. We do this in two steps:
\begin{itemize}
\item first showing that $\gamma_2(\hat{\Xi}^{(n)},\bar{d}^{(n)}) =\gamma_2(\hat{\Xi}_n,d)$ 
; and 
\item then showing $\gamma_2(\hat{\Xi}_n,d) \leq c_{4}w(\hat{\Xi}_n)$.
\end{itemize}

For the first step, note that there is a correspondence between $\hat{\xi}^{(n)} \in \hat{\Xi}^{(n)}$ and $\hat{\xi}_n \in \hat{\Xi}_n$ as $\hat{\xi}_n$ is a scaled stacked version of the $n$-tuples. Moreover, for any pair $\hat{\xi}^{(n)}_1, \hat{\xi}^{(n)}_2 \in \hat{\Xi}^{(n)}$ respectively corresponding to $\hat{\xi}_{1,n}, \hat{\xi}_{2,n} \in \hat{\Xi}_n$, by definition of $\bar{d}^{(n)}$ as in \eqref{eq:ncdist} and $d$ as in \eqref{eq:canon2}, we have 
\begin{align*}
d(\hat{\xi}_{1,n}, \hat{\xi}_{2,n}) & = 
\E\left[ \left( X_{\hat{\xi}_{1,n}} - X_{\hat{\xi}_{2,n}} \right)^2 \right] \\
& = \frac{1}{n}\mathbb{E}\left[\left(\langle \hat{\xi}_{1,n},g\rangle -\langle \hat{\xi}_{2,n},g\rangle\right)^{2}\right]\\
&=\frac{1}{n}\mathbb{E}\left[ \langle \hat{\xi}_{1,n}-\hat{\xi}_{2,n},g\rangle^{2}\right]\\
& = \frac{1}{n} \sum_{i=1}^n \| \xi_{i,1} - \xi_{i,2} \|_2^2 \\
& = \bar{d}^{(n)}(\hat{\xi}^{(n)}_1, \hat{\xi}^{(n)}_2)~.
\end{align*}

Further, because of the correspondence between the elements of $\hat{\Xi}^{(n)}$ and $\hat{\Xi}_n$, there is also a correspondence between admissible sequences $\Gamma^{(n)}$ of $\hat{\Xi}^{(n)}$ and $\Gamma$ of $\hat{\Xi}_n$. As a result, we have
\begin{align}
 \gamma_2(\hat{\Xi}^{(n)},\bar{d}^{(n)}) & = \inf_{\Gamma^{(n)}} \sup_{\hat{\xi}^{(n)} \in \hat{\Xi}^{(n)}} \sum_{r=0}^{\infty} 2^{r/2} \bar{d}^{(n)}(\hat{\xi}^{(n)}, \Gamma^{(n)}_r) \\   
 & = \inf_{\Gamma} \sup_{\hat{\xi}_n \in \hat{\Xi}_n} \sum_{r=0}^{\infty} 2^{r/2} d(\hat{\xi}_n, \Gamma_r) \\
 & = \gamma_2(\hat{\Xi}_n,d)~.
\end{align}

For the second step, since for the Gaussian process $X_{\hat{\xi}_n} = \langle \hat{\xi}_n, \g \rangle$ on $\hat{\Xi}_n$, the canonical distance is
\begin{align}
 \left( \E \left[ \left( X_{\hat{\xi}_{1,n}} - X_{\hat{\xi}_{2,n}} \right)^2 \right] \right)^{1/2} = \left( \frac{1}{n} \sum_{i=1}^n \| \xi_{i,1} - \xi_{i,2} \|^2_2 \right)^{1/2}~,
\end{align}
from the majorizing measure theorem \cite{tala14}, for some constant $c_4$, we have
\begin{align}
\gamma_2(\hat{\Xi}_n,d) \leq c_4 w(\hat{\Xi}_n)
\end{align}
That completes the proof. \qed

\section{SGD with Sample Reuse: Proofs for Section~\ref{sec:sgd}}
\label{app:sgd}

In this section, we provide proofs for results in Section~\ref{sec:sgd}.




\subsection{GD with sample reuse}

\vecadagd*

\proof Remembering that the population gradient $\nabla \cL_{D}(\theta_t) = \E_{z \sim \cD}[\nabla \ell(\theta_t; z)]$, for $t \in [T]$
%
%
%
consider the sequence of events 
{\small
\begin{align}
\hspace*{-5mm}
\Lambda_0  \triangleq  \left\{ \sup_{\theta_0 \in \Theta_0} \bigg\| \frac{1}{n} \sum_{i=1}^n \nabla \ell(\theta_0,z_i) - \nabla \cL_{D}(\theta_0)  \bigg\|_2 \leq \epsilon \right\}~, &~~\text{and} ~~ \bar{\Lambda}_0  \triangleq \left\{ \sup_{\theta_0 \in \Theta_0}  \bigg\| \frac{1}{n} \sum_{i=1}^n \nabla \ell(\theta_0,z_i) - \nabla \cL_{D}(\theta_0)  \bigg\|_2 > \epsilon \right\}~, \label{eq:lambda_0} \\
\Lambda_t  \triangleq  \left\{ \bigg\| \frac{1}{n} \sum_{i=1}^n \nabla \ell(\theta_t,z_i) - \nabla \cL_{D}(\theta_t)  \bigg\|_2 \leq \epsilon \right\}~, &~~\text{and} ~~ \bar{\Lambda}_t  \triangleq \left\{  \bigg\| \frac{1}{n} \sum_{i=1}^n \nabla \ell(\theta_t,z_i) - \nabla \cL_{D}(\theta_t)  \bigg\|_2 > \epsilon \right\}~. \label{eq:lambda_t}
\end{align}}
We define a filtration based these sequence of events $\Lambda_t$ which considers conditional events $\lambda | \{ \Lambda_{\tau \leq t }\}$with corresponding conditional probability distributions:
\begin{align}
\cD_0 & = \cD~, \\
\cD_t & = \cD \mid \{ \Lambda_{\tau < t} \}~. 
\end{align}
To make the analysis precise, we use $\P_{\cD_{t}}[\lambda]$ to denote probability of events $\lambda$ in the $t$-th stage of the filtration and $\P_{\cD}[\lambda]$ to denote probability of suitable events $\lambda$ according to the original distribution $\cD$. 

Note that (posterior) probability of any event $\lambda_1$ at stage $1$, i.e., conditioned on $\Lambda_1$ of the filtration can be written in terms of probabilities at the previous stage based on the definition of conditional probability as
\begin{equation}
\P_{\cD_{1}}[\lambda_1] = \P_{\cD| \{ \Lambda_0 \}}[\lambda_1] = \frac{ \P_{\cD}[\lambda_1, \Lambda_0]}{\P_{\cD}[\Lambda_0]}~.
\end{equation}
More generally, by the definition of conditional probability
\begin{equation}
\P_{\cD_{t}}[\lambda] = \P_{\cD| \{ \Lambda_{\tau < t} \}}[\lambda_t] = \frac{ \P_{\cD}[\lambda_t, \{ \Lambda_{\tau < t} \}]}{\P_{\cD}[\{ \Lambda_{\tau < t} \}]}~,
\label{eq:bayes2}
\end{equation}
where $\{ \Lambda_{\tau < t} \}$ denotes the joint event $\{\Lambda_0,\Lambda_1,\ldots, \Lambda_{t-1}\}$ where individual events $\Lambda_{\tau}$ are as in \eqref{eq:lambda_0} and \eqref{eq:lambda_t}. While seemingly straightforward, \eqref{eq:bayes2} shows a way to express probabilities $\P_{\cD_{t}}[\cdot]$ of events in the sequence directly in terms $\P_{\cD}[\cdot]$, the original distribution. 

For our analysis, instead of working with the full distributions, we will be working on the finite sample versions based on $z^{(n)} \sim \cD^n$. In particular, let $\cD_0(z^{(n)})$ be the empirical distribution based on the samples, i.e., $\cD_0(z^{(n)}) = \frac{1}{n}\sum_{i=1}^n \1_{z_i}$, where $\1$ is the indicator function. As before, the
(posterior) probability of any event $\lambda_1$ at any stage $1$ of the filtration can be written in terms of probabilities at the previous stage based on the definition of conditional probability:
\begin{equation}
\P_{\cD_{1}(z^{(n)})}[\lambda_1] = \P_{\cD_0(z^{(n)})| \{ \Lambda_0 \}}[\lambda_1] = \frac{ \P_{\cD_0(z^{(n)})}[\lambda_1, \Lambda_0]}{\P_{\cD_0(z^{(n)})}[\Lambda_0]}~.
\end{equation}
More generally, by the definition of conditional probability:
\begin{equation}
\P_{\cD_{t}(z^{(n)})}[\lambda_t] = \P_{\cD_0(z^{(n)})| \{ \Lambda_{\tau \leq t} \}}[\lambda_t] = \frac{ \P_{\cD_0(z^{(n)})}[\lambda_t, \{ \Lambda_{\tau < t} \}]}{\P_{\cD_0(z^{(n)})}[\{ \Lambda_{\tau < t} \}]}~.
\label{eq:bayes3}
\end{equation}
Note that events, $\Lambda_t$, of interest are defined in terms of the samples $z^{(n)}$, and the definition of conditional probability provides a way to express the probabilities of such events under $\cD_{t}(z^{(n)})$ in terms of events under $\cD_{0}(z^{(n)})$. Since  $\cD_{0}(z^{(n)})$ is the empirical distribution of the samples $z^{(n)} \sim \cD^n$, for convenience, we will denote $\P_{\cD_{0}(z^{(n)})}[\cdot]$ as $\P_{z^{(n)} \sim \cD^n}[\cdot]$ and even more briefly as $\P_{z^{(n)}}[\cdot]$.



To simplify the notation, consider the random variable 
\begin{align}
{X}_{0} & \triangleq \sup_{\theta_0 \in \Theta_0} \bigg\| \frac{1}{n} \sum_{i=1}^n \nabla \ell(\theta_0,z_i) - \nabla \cL_{D}(\theta_0)  \bigg\|_2 ~,\\
{X}_{t} & \triangleq \bigg\| \frac{1}{n} \sum_{i=1}^n \nabla \ell(\theta_t,z_i) - \nabla \cL_{D}(\theta_t)  \bigg\|_2 ~.
\end{align}
Then, from \eqref{eq:lambda_0} and \eqref{eq:lambda_t}, we have $\Lambda_t = \{ X_t \leq \epsilon\}$ and $\bar \Lambda_t = \{ X_t > \epsilon\}$. Then, we have 
\begin{align}
\P & \left[ \max \left(  \sup_{\theta_0 \in \Theta_0}  \left\| \frac{1}{n} \sum_{i=1}^n \nabla \ell(\theta_0,z_i) - \nabla \cL_D(\theta_0) \right\|_2 ~,~~~\max_{t \in [T]} \bigg\| \frac{1}{n} \sum_{i=1}^n \nabla \ell(\theta_t,z_i) - \nabla \cL_{D}(\theta_t)  \bigg\|_2 \right) > \epsilon\right] \\
& \leq \P_{z^{(n)}} \left[ \sup_{\theta_0 \in \Theta_0} \bigg\| \frac{1}{n} \sum_{i=1}^n \nabla \ell(\theta_0,z_i) - \nabla \cL_{D}(\theta_0)  \bigg\|_2 > \epsilon \right] +  \sum_{t = 1}^{T} \P_{\cD_{t}(z^{(n)})} \left[ \bigg\| \frac{1}{n} \sum_{i=1}^n \nabla \ell(\theta_t,z_i) - \nabla \cL_{D}(\theta_t)  \bigg\|_2 > \epsilon \right] \nonumber \\
& = \sum_{t=0}^{T} \P_{\cD_{t}(z^{(n)})}\left[ X_t > \epsilon \right] \nonumber  \\
& = \sum_{t=0}^{T} \P_{\cD_{t}(z^{(n)})}[\bar \Lambda_t] ~.
\end{align}
Now, with $\{\Lambda_{0:\tau}\} := (\Lambda_0,\Lambda_1,\ldots,\Lambda_{\tau})$ and $\{0:\tau\} := \{0,1,\ldots,\tau\}$, we have
\begin{align}
\sum_{t=0}^{T}  \P_{\cD_{t}(z^{(n)})}[\bar \Lambda_t] 
& = \P_{\cD_0(z^{(n)})}[\bar{\Lambda}_0] + \P_{\cD_1(z^{(n)})}[\bar{\Lambda}_1] + \ldots + \P_{\cD_{T}(z^{(n)})}[\bar{\Lambda}_T] \nonumber \\
& = \P_{z^{(n)}}[\bar{\Lambda}_0] + \frac{\P_{z^{(n)}}[\Lambda_0,\bar{\Lambda}_1]}{\P_{z^{(n)}}[\Lambda_0]} + \ldots + \frac{\P_{z^{(n)}}[\{\Lambda_{0:(T - 1)}\},\bar{\Lambda}_T]}{\P_{z^{(n)}}[\{\Lambda_{0:(T - 1)}\}]} \nonumber \\
& \overset{(a)}{\leq}  \frac{\P_{z^{(n)}}[\bar{\Lambda}_0]}{\P_{z^{(n)}}[\{\Lambda_{0:(T - 1)}\}]}  + \frac{\P_{z^{(n)}}[\Lambda_0,\bar{\Lambda}_1]}{\P_{z^{(n)}}[\{\Lambda_{0:(T - 1)}\}]}  + \ldots + \frac{\P_{z^{(n)}}[\{\Lambda_{0:(T - 1)}\},\bar{\Lambda}_T]}{\P_{z^{(n)}}[\{\Lambda_{0:(T - 1)}\}]} \nonumber \\
& \overset{(b)}{\leq} \frac{ 1 - \P_{z^{(n)}}[\{\Lambda_{0:T}\}]}{\P_{z^{(n)}}[\{\Lambda_{0:(T - 1)}\}]} \nonumber \\
& = \frac{1 - \P_{z^{(n)}} \left[ \max_{t \in \{0:T\}} X_t \leq \epsilon \right]}{\P_{z^{(n)}} \left[ \max_{t \in \{0:T-1\}} X_t \leq \epsilon\right]} \nonumber \\
\nonumber & = \frac{\P_{z^{(n)}} \left[ \max_{t \in \{0:T\}} X_t > \epsilon\right]}{\P_{z^{(n)}} \left[ \max_{t \in \{0:T-1\}} X_t \leq \epsilon \right]} \\
& \overset{(c)}{\leq} \frac{\P_{z^{(n)}} \left[ \max_{t \in \{0:T\}} X_t > \epsilon \right]}{\P_{z^{(n)}} \left[ \max_{t \in \{0:T\}} X_t \leq \epsilon \right]}~.\label{eq:transfer}
\end{align}
where (a) follows since $\P_{z^{(n)}}[\{\Lambda_{0:(T - 1)}\}] \leq \P_{z^{(n)}}[\{\Lambda_{0:\tau}\}]\leq1$ for all $\tau \in \{0:T\}$, (b) follows since the counter-event of $\{\Lambda_{0:T}\}$ covers the events in $(a)$,
and (c) follows since $\P_{z^{(n)}} \left[ \max_{t \in \{0:T\}} X_t \leq \epsilon \right] \leq \P_{z^{(n)}} \left[ \max_{t \in \{0:T-1\}} X_t \leq \epsilon \right]$.
Thus, it suffices to focus on an upper bound on 
$\P_{z^{(n)} \sim \cD^n} [ \max_{t \in \{0:T\}} X_t  > \epsilon]$
which corresponds to the same `bad' event 
$\{  \max_{t \in \{0:T\}} X_t  > \epsilon \}$
but in the non-adaptive setting, since the probability is w.r.t.~$\P_{z^{(n)} \sim \cD^n} \left[ \cdot \right]$. 
%

From Lemma \ref{lemm: gradient_union_bound}, 
for any $u>0$, we have
\begin{align}
    \P_{z^{(n)}} \left[ \sup_{\theta_0 \in \Theta_0} \bigg\| \frac{1}{n} \sum_{i=1}^n \nabla \ell(\theta,z_i) - \nabla \cL_{\cD}(\theta)   \bigg\|_2 >  \frac{c_1 w(\hat{\Xi}^0_n) + u}{\sqrt{n}}  \right] \leq  \exp(- 2u^2)~.
    \label{eq: gradient_concentration_bound}
\end{align}

Further, from Lemma \ref{lemm:vector_bernstein}, 
for any $u>0$, we have
\begin{align}
    \P_{z^{(n)}} \left[ \bigg\| \frac{1}{n} \sum_{i=1}^n \nabla \ell(\theta,z_i) - \nabla \cL_{\cD}(\theta)   \bigg\|_2 >  \frac{c_1 \left(\log p + u\right)}{\sqrt{n}}  \right] \leq  \exp(- u^{2})~.
    \label{eq: gradient_norm_bound}
\end{align}

Choosing $\epsilon = \frac{c_1 \max(w(\hat{\Xi}_n),\log p) + u}{\sqrt{n}}$ and $u = \log (T+1) + \log \frac{1}{\delta}$, we have
\begin{align*}
    \P_{z^{(n)}} \left[ \max_{t \in \{0:T\}} X_t > \epsilon \right] & \overset{(a)}{\leq}  \sum_{t=0}^T\P_{z^{(n)}} \left[  X_t > \epsilon \right] =\mathbb{P}_{z^{(n)}}\left[X_{0}>\epsilon\right]+\sum_{t=1}^{T}\mathbb{P}_{z^{(n)}}\left[X_{t}>\epsilon\right]\\
& \overset{(b)}{\leq} \exp(-2u^2) + T \exp(-u^{2}) \\
& \overset{(c)}{\leq} (T+1) \exp(-u^{2}) \\
& \overset{(d)}{\leq} (T+1)  \exp\left(-\left(\log (T+1) + \log \frac{1}{\delta}\right)\right) 
     =\delta,
\end{align*}
where $(a)$ follows by union bound, (b) follows from \ref{eq: gradient_concentration_bound} and \ref{eq: gradient_norm_bound}, (c) follows since $u > 1$ for $T \geq 2$, and (d) follows from the form of $u$. Then, for $\delta \leq 1/2$
we have 
\begin{align*}
\P_{\cD_T(z^{(n)})} \left[ \max_{t \in \left\{0:T\right\}} X_t  > \epsilon \right] & \leq \frac{\P_{z^{(n)}} \left[ \max_{t \in \left\{0:T\right\}} X_t  > \epsilon  \right]}{\P_{z^{(n)}} \left[ \max_{t \in \left\{0:T\right\}} X_t  \leq \epsilon  \right]}  \leq \frac{\delta}{1-\delta} \leq 2 \delta~.
\end{align*}
That completes the proof. \qed

\begin{restatable}{lemm}{gradient_union_bound}
\label{lemm: gradient_union_bound}
With $\hat \Xi_n,  w(\hat \Xi_n) $ respectively denoting the set of empirical gradients and its Gaussian width as in Definition~\ref{defn:gaussw}, for any $u > 0$, we have
\begin{align}
    \P_{z^{(n)}} \left[ \sup_{\theta \in \Theta} \bigg\| \frac{1}{n} \sum_{i=1}^n \nabla \ell(\theta,z_i) - \nabla \cL_{\cD}(\theta)   \bigg\|_2 >  \frac{c_1 w(\hat{\Xi}_n) + u}{\sqrt{n}}  \right] \leq  \exp(- 2u^2)~.
\end{align}
\end{restatable}
\proof From \cite{FSS}, with $\nabla \hat{\cL}_n(\theta) = \frac{1}{n} \sum_{i=1}^n \nabla \ell(\theta,z_i)$,  for any $\delta >0$, with probability at least $1-\delta$ over the data $z^{(n)}$, we have
\begin{align}
      \E_{z^{(n)}} \sup_{\theta \in \Theta}\| \nabla \hat{\cL}_n(\theta) -\nabla \cL_{\cD}(\theta) \| &\leq 4 \hat{R}_n(\hat{\Xi}^{(n)}) +  \frac{\log \frac{1}{\delta}}{n}~.
\end{align}
Using the bounded difference inequality~\cite{boucheron_concentration_2013},  for any $\epsilon >0$ and $\|\nabla \ell(\theta;z)\|_2 = O(1)$ for $\theta \in \Theta, z \in \cZ$, we have 
\begin{align}
    \P_{z^{(n)}} \left[ \sup_{\theta \in \Theta} \bigg\| \frac{1}{n} \sum_{i=1}^n \nabla \ell(\theta,z_i) - \nabla \cL_{\cD}(\theta)   \bigg\|_2 >  \E_{z^{(n)}} \sup_{\theta \in \Theta}\| \nabla \hat{\cL}_n(\theta) -\nabla \cL_{\cD}(\theta) \| + \epsilon  \right] \leq  \exp(- 2n\epsilon^2)~.
\end{align}

From Section \ref{sec:gen} and Theorem \ref{theo:rcgc}, we have
\begin{align}
         \hat{R}_n(\hat{\Xi}^{(n)}) \leq c_2 \frac{w(\hat{\Xi}_n)}{\sqrt{n}}~.
\end{align}

Combining the above results, with $\epsilon = \frac{u}{\sqrt{n}}$, for any $\delta >0$, with probability at least $1-\delta$ over the data $z^{(n)}$, we have
\begin{align}
    \P_{z^{(n)}} \left[ \sup_{\theta \in \Theta} \bigg\| \frac{1}{n} \sum_{i=1}^n \nabla \ell(\theta,z_i) - \nabla \cL_{\cD}(\theta)   \bigg\|_2 >  \frac{c_1 w(\hat{\Xi}_n) + u}{\sqrt{n}}  \right] \leq  \exp(- 2u^2)~,
\end{align}
where $c_1 = 4 c_2$. That completes the proof. \qed

\begin{lemm}[\bf Matrix Bernstein Inequality]\cite{tropp_matrix_concentration_2015}\label{lemma:matrix_bernstein}
Let $S_{1}$,...,$S_{n}$ be independent, centered random matrices with common dimension $d_{1}\times d_{2}$, and assume that each one is uniformly bounded
\begin{align*}
    \mathbb{E}S_{k}=0, \left\|S_{k}\right\|\leq L, \forall k\in[n].
\end{align*}
Introduce the sum
\begin{align*}
    Z=\sum_{k=1}^{n}S_{k},
\end{align*}
and let $v(Z)$ denote the matrix variance statistic of the sum
\begin{align*}
    v(Z)&=\max\left\{\left\|\mathbb{E}\left(ZZ^{*}\right)\right\|,\left\|\mathbb{E}\left(Z^{*}Z\right)\right\|\right\}\\
    &=\max\left\{\left\|\sum_{k=1}^{n}\mathbb{E}\left(S_{k}S_{k}^{*}\right)\right\|,\left\|\sum_{k=1}^{n}\mathbb{E}\left(S_{k}^{*}S_{k}\right)\right\|\right\}.
\end{align*}
Then
\begin{align*}
    \mathbb{P}\left\{\left\|Z\right\|\geq t\right\}\leq\left(d_{1}+d_{2}\right)\cdot\exp\left(\frac{-t^{2}/2}{v(Z)+Lt/3}\right), \forall t\geq0.
\end{align*}
Furthermore,
\begin{align*}
    \mathbb{E}\left\|Z\right\|\leq\sqrt{2v(Z)\log(d_{1}+d_{2})}+\frac{1}{3}L\log(d_{1}+d_{2}).
\end{align*}
\end{lemm}

\begin{restatable}{lemm}{vector_bernstein}
\label{lemm:vector_bernstein}
Under Assumption~\ref{asmp:gradbnd}, for any $u>0$, we have
\begin{align}
    \P_{z^{(n)}} \left[ \bigg\| \frac{1}{n} \sum_{i=1}^n \nabla \ell(\theta,z_i) - \nabla \cL_{\cD}(\theta)   \bigg\|_2 >  \frac{c_1 \left(\log p + u\right)}{\sqrt{n}}  \right] \leq  \exp(- u^{2})~.
\end{align}
\end{restatable}
\proof Following Assumption~\ref{asmp:gradbnd}, denote the $L_{2}$ norm bound of the gradient is $G$. There are two different cases:
\begin{itemize}
    \item If $\log p+u\geq\sqrt{n}$, then applying the bounded difference inequality~\cite{boucheron_concentration_2013}, for any $\epsilon>0$, 
    \begin{align*}
        \mathbb{P}_{z^{(n)}}\left[\left\|\frac{1}{n}\sum_{i=1}^{n}\nabla\ell\left(\theta,z_{i}\right)-\nabla\mathcal{L}_{\mathcal{D}}\left(\theta\right)\right\|_{2}>\mathbb{E}_{z^{(n)}}\left\|\nabla\hat{\mathcal{L}}_{n}\left(\theta\right)-\nabla\mathcal{L}_{\mathcal{D}}\left(\theta\right)\right\|+\epsilon\right]\leq\exp\left(-\frac{n\epsilon^{2}}{2G^{2}}\right)
    \end{align*}
    Since
    \begin{align*}
        \mathbb{E}_{z^{(n)}}\left\|\nabla\hat{\mathcal{L}}_{n}\left(\theta\right)-\nabla\mathcal{L}_{\mathcal{D}}\left(\theta\right)\right\|\leq 2G=\frac{2\sqrt{n}G}{\sqrt{n}}\leq\frac{2G\left(\log p+u\right)}{\sqrt{n}}
    \end{align*}
    then taking $\epsilon = \frac{\sqrt{2}Gu}{\sqrt{n}}$, we have
    \begin{align*}
        \mathbb{P}_{z^{(n)}}&\left[\left\|\frac{1}{n}\sum_{i=1}^{n}\nabla\ell\left(\theta,z_{i}\right)-\nabla\mathcal{L}_{\mathcal{D}}\left(\theta\right)\right\|_{2}>\frac{2G\left(\log p+u\right)+\sqrt{2}Gu}{\sqrt{n}}\right]\\
        &\leq\mathbb{P}_{z^{(n)}}\left[\left\|\frac{1}{n}\sum_{i=1}^{n}\nabla\ell\left(\theta,z_{i}\right)-\nabla\mathcal{L}_{\mathcal{D}}\left(\theta\right)\right\|_{2}>\mathbb{E}_{z^{(n)}}\left\|\nabla\hat{\mathcal{L}}_{n}\left(\theta\right)-\nabla\mathcal{L}_{\mathcal{D}}\left(\theta\right)\right\|+\frac{\sqrt{2}Gu}{\sqrt{n}}\right]\\
        &\leq\exp(-u^{2})
    \end{align*}
    Therefore,
    \begin{align*}
        \mathbb{P}_{z^{(n)}}\left[\left\|\frac{1}{n}\sum_{i=1}^{n}\nabla\ell\left(\theta,z_{i}\right)-\nabla\mathcal{L}_{\mathcal{D}}\left(\theta\right)\right\|_{2}>\frac{2G\left(\log p+u\right)+\sqrt{2}Gu}{\sqrt{n}}\right]\leq\exp(-u^{2})
    \end{align*}
    \item If $\log p+u<\sqrt{n}$, then with $S_{i}=\nabla\ell\left(\theta_{t},z_{i}\right)-\mathbb{E}\nabla\ell\left(\theta_{t},z_{i}\right)$ and $Z=\sum_{i=1}^{n}S_{i}$, we have $\mathbb{E}S_{i}=0$, and 
\begin{align*}\left\|S_{i}\right\|=\left\|\nabla\ell\left(\theta_{t},z_{i}\right)-\mathbb{E}\nabla\ell\left(\theta_{t},z_{i}\right)\right\|\leq\left\|\nabla\ell\left(\theta_{t},z_{i}\right)\right\|+\left\|\mathbb{E}\nabla\ell\left(\theta_{t},z_{i}\right)\right\|\leq2G.
\end{align*}
so we can take $L=2G$ as an upper bound of $\left\|S_{i}\right\|$. Besides,
\begin{align*}
    v(Z)=\max\left\{\left\|\sum_{i=1}^{n}\mathbb{E}\left(S_{i}S_{i}^{*}\right)\right\|,\left\|\sum_{i=1}^{n}\mathbb{E}\left(S_{i}^{*}S_{i}\right)\right\|\right\}\leq n\max_{i\in[n]}\left\|S_{i}\right\|^{2}\leq4nG^{2}
\end{align*}
Then directly apply Lemma \ref{lemma:matrix_bernstein} with $d_{1}=1$, $d_{2}=p$, we can get that
\begin{align*}
    \mathbb{P}\left\{\left\|Z\right\|\geq t\right\}\leq\left(p+1\right)\cdot\exp\left(-\frac{t^{2}/2}{v(Z)+Lt/3}\right)
\end{align*}
Substituting $L=2G$ and $v(Z)\leq 4nG^{2}$, we have
\begin{align*}
    \mathbb{P}\left\{\left\|Z\right\|\geq t\right\}\leq\left(p+1\right)\cdot\exp\left(-\frac{t^{2}/2}{4nG^{2}+2Gt/3}\right)=\left(p+1\right)\cdot\exp\left(\frac{-3t^{2}}{24nG^{2}+4Gt}\right)
\end{align*}
Taking $t=4\sqrt{n}G\left(\log p+u\right)$, we have
\begin{align*}
    \mathbb{P}\left\{\left\|Z\right\|\geq4\sqrt{n}G\left(\log p+u\right)\right\}&\leq\left(p+1\right)\cdot\exp\left(-\frac{48nG^{2}\left(\log p+u\right)^{2}}{24nG^{2}+16\sqrt{n}G^{2}\left(\log p+u\right)}\right)\\
    &<\left(p+1\right)\cdot\exp\left(-\frac{48nG^{2}\left(\log p+u\right)^{2}}{24nG^{2}+16\sqrt{n}G^{2}\cdot\sqrt{n}}\right)\\
    &<\left(p+1\right)\cdot\exp\left(-\left(\log p+u\right)^{2}\right)<\left(p+1\right)\cdot\exp\left(-\left(\left(\log p\right)^{2}+u^{2}\right)\right)\\
    &\leq\exp\left(-u^{2}\right)
\end{align*}
So we have
\begin{align*}
        \mathbb{P}_{z^{(n)}}&\left[\left\|\frac{1}{n}\sum_{i=1}^{n}\nabla\ell\left(\theta,z_{i}\right)-\nabla\mathcal{L}_{\mathcal{D}}\left(\theta\right)\right\|_{2}>\frac{4G\left(\log p+u\right)}{\sqrt{n}}\right]\\
        &=\mathbb{P}\left\{\frac{1}{n}\left\|Z\right\|\geq\frac{4G\left(\log p+u\right)}{\sqrt{n}}\right\}=\mathbb{P}\left\{\left\|Z\right\|\geq4\sqrt{n}G\left(\log p+u\right)\right\}\\
        &\leq\exp(-u^{2})
    \end{align*}
    Therefore,
    \begin{align*}
        \mathbb{P}_{z^{(n)}}&\left[\left\|\frac{1}{n}\sum_{i=1}^{n}\nabla\ell\left(\theta,z_{i}\right)-\nabla\mathcal{L}_{\mathcal{D}}\left(\theta\right)\right\|_{2}>\frac{4G\left(\log p+u\right)}{\sqrt{n}}\right]\leq\exp(-u^{2})
    \end{align*}
\end{itemize}
Combining these two cases, there exist $c_{3}$ such that
\begin{align*}
    \mathbb{P}_{z^{(n)}}&\left[\left\|\frac{1}{n}\sum_{i=1}^{n}\nabla\ell\left(\theta,z_{i}\right)-\nabla\mathcal{L}_{\mathcal{D}}\left(\theta\right)\right\|_{2}>\frac{c_{3}G\left(\log p+u\right)}{\sqrt{n}}\right]\leq\exp(-u^{2})
\end{align*}
Since $G=O(1)$ according to Assumption \ref{asmp:gradbnd}, by taking $c_{1}=Gc_{3}$, we complete the proof.
\subsection{Population Convergence in Optimization}

\theosgduc*

\proof 
%
With the condition that $\cL_{\cD}(\theta)$ has $\tau$-Lipschitz gradient, i.e., for any $\theta, \theta^\prime \in \Theta$ 
\begin{align}
    \|\nabla \cL_{\cD}(\theta) - \nabla \cL_{\cD}(\theta^\prime)\|_2 \leq  \tau \| \theta - \theta^\prime \|,
\end{align}
we have
\begin{align}
\cL_{\cD}(\theta) \leq \cL_{\cD}(\theta^\prime) + \langle \theta -\theta^\prime, \cL_{\cD}(\theta) \rangle + \frac{\tau}{2}\cdot \| \theta -\theta^\prime\|_2^2.
\end{align}
At iteration $t$, to simplify the notation, we use the $g(\theta_t) = \frac{1}{n} \sum_{i=1}^{n} \nabla \ell\left(\theta_{t}, z_{i}\right)$ to be the sample gradient estimate and $\nabla \cL_{\cD}(\theta_t)$ is the population gradient. Let $\Delta_t = g(\theta_t) - \nabla \cL_{\cD}(\theta_t)$. 

Given the update of SGD to be $\theta_{t+1} = \theta_{t} -\eta_t g(\theta_t)$,
we have
\begin{align} \label{eq:decent}
\nr \cL_{\cD}(\theta_{t+1}) &\leq \cL_{\cD}(\theta_t) + \langle \nabla \cL_{\cD}(\theta_t), \theta_{t+1} -\theta_t \rangle + \frac{\tau \eta_t^2}{2} \|\theta_{t+1} -\theta_t \|_2^2 \\
\nr & \leq  \cL_{\cD}(\theta_t) - \eta_t \langle \nabla \cL_{\cD}(\theta_t),  \nabla \cL_{\cD}(\theta_t) +  \Delta_t \rangle + \frac{\tau \eta_t^2}{2}\| \nabla \cL_{\cD}(\theta_t) + \Delta_t \|_2^2  \\
\nr &\leq  \cL_{\cD}(\theta_t) - \frac{\eta_t}{2} \| \nabla \cL_{\cD}(\theta_t)\|_2^2  + \frac{\eta_t}{2} \| \Delta_t\| _2^2 + \tau \eta_t^2 \|\nabla \cL_{\cD}(\theta_t) \|^2  + \tau \eta_t^2 \|\Delta_t \|_2^2 \\
& =  \cL_{\cD}(\theta_t) - \left(\frac{\eta_t}{2} - \tau \eta_t^2 \right) \| \nabla \cL_{\cD}(\theta_t)\|_2^2 + \left(\frac{\eta_t}{2} + \tau \eta_t^2 \right) \|\Delta_t \|_2^2 .
\end{align}

From Theorem \ref{thm:vecada}, with probability at least $1-2\delta$, for all $t \in \{0:T\}$
we have 
\begin{equation}
    \| \Delta_t \|_2^2 \leq \frac{\left(c_1 \max(w(\hat{\Xi}^0_{n}),\log p ) ) + \sqrt{\log T + \log \frac{1}{\delta}}\right)^{2}}{n}\leq\frac{2c_{1}^{2}\max\left(w^{2}(\hat{\Xi}^{0}_{n}),\log^{2}p\right)+2\left(\log T+\log\frac{1}{\delta}\right)}{n}.
\end{equation}

Bring this to the \eqref{eq:decent} and sum over iteration from $t=1$ to $t = T$, we have

\begin{equation}
\frac{1}{T}\sum_{t=1}^{T} \left(\frac{\eta_t}{2} - \tau \eta_t^2 \right)\| \nabla \cL_{\cD}(\theta_t)\|_2^2  \leq \frac{\cL_{\cD}(\theta_t) - L^\star_{\cD}}{ T} + \frac{1}{T}\sum_{t=1}^{T}\left(\frac{\eta_t}{2} + \tau \eta_t^2 \right) \|\Delta_t \|_2^2,
\end{equation}
where $\cL_{\cD}(\theta_{T+1})$ is lower bounded by $L^\star_{\cD}$. 
Choosing $\eta_t = \frac{1}{4\tau}$, we have
\begin{align}
\frac{1}{T}\sum_{t=1}^{T} \| \nabla \cL_{\cD}(\theta_t)\|_2^2  & \leq \frac{16\tau\left( \cL_{\cD}(\theta_T) - L^\star_{\cD}\right)}{T} + \frac{1}{T}\sum_{t=1}^{T}3 \|\Delta_t \|_2^2  \\
& \leq \frac{16\tau\left( \cL_{\cD}(\theta_T) - L^\star_{\cD}\right)}{ T}+ 3\frac{2c_{1}^{2}\max\left(w^{2}(\hat{\Xi}^{0}_{n}),\log^{2}p\right)+2\left(\log T+\log\frac{1}{\delta}\right)}{n}
\end{align}
Given that $\theta_R$ is uniformly sampled from $\{\theta_1, ..., \theta_T \}$, we have
\begin{equation}
    \mathbb{E}[ \| \nabla L_\cD (\theta_R)\|_2^2] = \frac{1}{T}\sum_{t=1}^{T}\| \nabla \cL_{\cD}(\theta_t)\|_2^2
\end{equation}

That completes the proof. \qed

\section{Gaussian Width of Gradients: Proofs for Section~\ref{sec:rad_grad}}
\label{app:gauss_width}
\subsection{Gaussian Width bounds for Feed-Forward Networks (FFNs)}
\label{app:ffn}
We consider $f$ to be a FFN with given by
\begin{equation}
f(\theta;\x) = \v^\top \phi(\frac{1}{\sqrt{m_{L}}}W^{(L)} \phi(\cdots \phi(\frac{1}{\sqrt{m_{1}}}  W^{(1)} \x))))~,
\label{eq:ffn_app}    
\end{equation}
where $W^{(1)}\in\mathbb{R}^{m_{1}\times d}$, $W^{(l)}\in\mathbb{R}^{m_{l}\times m_{l-1}},l\in\left\{2,...,L\right\}$ are layer-wise weight matrices, $\v\in\mathbb{R}^{m_{L}}$ is the last layer vector, $\phi(\cdot)$ is the smooth (pointwise) activation function, and the total set of parameters
\begin{align}
    \theta:=\left(\text{vec}\left(W^{(1)}\right)^{\top},...,\text{vec}\left(W^{(L)}\right)^{\top},v^{\top}\right)^{\top}\in\mathbb{R}^{\sum_{k=1}^{L}m_{k}m_{k-1}+m_{L}}
    \label{eq:theta_def_1}
\end{align}
with $m_{0}=d$. For convenience, we write the model in terms of the layerwise outputs or features as:
\begin{align}
\alpha^{(0)}(\x) & = \x~, \\
\alpha^{(l)} & = \phi\left( \frac{1}{\sqrt{m_{l}}} W^{(l)} \alpha^{(l-1)}(\x) \right) \\
f(\theta;\x) & = \v^\top \alpha^{(L)}(\x)~.
\end{align}
Our analysis for both FFNs and ResNets will be for gradients over all parameters in a fixed radius spectral norm ball around the initialization $\theta_0$:
\begin{equation}
\begin{split}
B_{\rho, \rho_1}^{\spec}(\theta_0) := &\big\{ \theta \in \R^p ~\text{as in \eqref{eq:theta_def_1}} ~\mid  \| \v - \v_0 \|_2 \leq \rho_1,~   \|W^{(\ell)} - W_0^{(\ell)} \|_2 \leq \rho, \ell \in [L]  \big\} 
\label{eq:specball_app} ~.
\end{split}
\end{equation}
As is standard, we assume $\|\x\|_2 =1$ according to Assumption~\ref{asmp:ginit}.

We start with the following standard consequence of the Gaussian random initialization, variants of which is widely used in practice~\cite{allen-zhu_convergence_2019,du_gradient_2019,arora2019fine,banerjee_rsc23}.
\begin{lemm}\cite{banerjee_rsc23}\label{lemma:matrix}
Under Assumption~\ref{asmp:ginit}, for $\theta\in B_{\rho,\rho_{1}}^{\text{Spec}}(\theta_{0})$, with probability at least $1-\frac{2}{m_{l}}$, 
\begin{align*}
    \left\|W^{(l)}\right\|_{2}\leq\left(\sigma_{1}+\frac{\rho}{\sqrt{m_{l}}}\right)\sqrt{m_{l}}=\beta_{l}\sqrt{m_{l}}
\end{align*}
with $\beta_{l}=\sigma_{1}+\frac{\rho}{\sqrt{m_{l}}}$.
\end{lemm}
\proof For a ($m_{l}\times m_{l-1}$) random matrix $W_{0}^{(l)}$ with i.i.d. entries $w_{0,ij}^{(l)}\sim\mathcal{N}(0,\sigma_{0}^{(l)}{}^{2})$, with probability at least $1-2\exp(-t^{2}/2\sigma_{0}^{(l)}{}^{2})$, the largest singular value of $W_{0}$ is bounded by
\begin{align*}
    \sigma_{\max}\left(W_{0}^{(l)}\right)\leq\sqrt{m_{l}}+\sqrt{m_{l-1}}+t
\end{align*}
Let us choose $t=\sigma_{0}^{(l)}\sqrt{2\log m_{l}}$ so that the inequality holds with probability at least $1-\frac{2}{m_{l}}$. Then, there are two cases:
\begin{itemize}
    \item \textbf{Case 1:} $l=1$. Since $m_{0}=d$ and $m_{1}\geq d$, with probability at least $1-\frac{2}{m_{1}}$,
    \begin{align*}
        \left\|W_{0}^{(1)}\right\|_{2}\leq\sigma_{0}^{(1)}\left(\sqrt{m_{1}}+\sqrt{d}+\sqrt{2\log m_{1}}\right)\leq\sigma_{0}^{(1)}\left(2\sqrt{m_{1}}+\sqrt{2\log m_{1}}\right)=\sigma_{1}\sqrt{m_{1}}
    \end{align*}
    \item \textbf{Case 2:} $2\leq l\leq L$. With probability at least $1-\frac{2}{m_{l}}$, 
    \begin{align*}
        \left\|W_{0}^{(l)}\right\|_{2}\leq\sigma_{0}^{(l)}\left(\sqrt{m_{l}}+\sqrt{m_{l-1}}+\sqrt{2\log m_{l}}\right)=\sigma_{1}\sqrt{m_{l}}
    \end{align*}
\end{itemize}
Then, by triangle inequality, for $\theta\in B_{\rho,\rho_{1}}^{\text{Spec}}\left(\theta_{0}\right)$, 
\begin{align*}
    \left\|W^{(l)}\right\|_{2}\leq\left\|W_{0}^{(l)}\right\|_{2}+\left\|W^{(l)}-W_{0}^{(l)}\right\|_{2}\overset{(a)}{\leq}\sigma_{1}\sqrt{m_{l}}+\rho=\beta_{l}\sqrt{m_{l}}
\end{align*}
\begin{lemm}
\label{lemma:layer}
Consider any $l\in\left[L\right]$. Under Assumption~\ref{asmp:act} and \ref{asmp:ginit}, for $\theta\in B_{\rho,\rho_{1}}^{\text{Spec}}\left(\theta_{0}\right)$, with probability at least $1-\sum_{k=1}^{l}\frac{2}{m_{k}}$, we have
\begin{align*}
    \left\|\alpha^{(l)}\right\|_{2}\leq \prod_{k=1}^{l}\left(\sigma_{1}+\frac{\rho}{\sqrt{m_{k}}}\right) =\prod_{k=1}^{l}\beta_{k}
\end{align*}
with $\beta_{k}=\sigma_{1}+\frac{\rho}{\sqrt{m_{k}}}$.
\end{lemm}
\proof Following~\cite{allen-zhu_convergence_2019,CL-LZ-MB:20}, we prove the result by recursion. First, recall that since $\| \x\|_2^2 = 1$, we have $\| \alpha^{(0)}\|_2 = 1$. Then, since $\phi$ is 1-Lipschitz and $\phi(0)=0$,
\begin{align*}
\left\|\phi\left( \frac{1}{\sqrt{m_{1}}} W^{(1)} \alpha^{(0)} \right) \right\|_2 - \| \phi(\mathbf{0}) \|_2 
\leq \left\|\phi\left( \frac{1}{\sqrt{m_{1}}} W^{(1)} \alpha^{(0)} \right) - \phi(\mathbf{0}) \right\|_2 \leq \left\|  \frac{1}{\sqrt{m_{1}}} W^{(1)} \alpha^{(0)} \right\|_2 ~,
\end{align*}
so that
\begin{align*}
\| \alpha^{(1)}\|_2 & = \left\| \phi\left( \frac{1}{\sqrt{m_{1}}} W^{(1)} \alpha^{(0)} \right) \right\|_2
 \leq  \left\|  \frac{1}{\sqrt{m_{1}}} W^{(1)} \alpha^{(0)} \right\|_2 + \| \phi(\mathbf{0}) \|_2\\ 
&\leq  \frac{1}{\sqrt{m_{1}}} \| W^{(1)} \|_2 \|\alpha^{(0)} \|_2  \\
& \leq \left( \sigma_1 + \frac{\rho}{\sqrt{m_{1}}}\right) ~,
\end{align*}
where we used Lemma \ref{lemma:matrix} in the last inequality, which holds with probability at least $1-\frac{2}{m_{1}}$. For the inductive step, we assume that for some $l-1$, we have 
\begin{align*}
\| \alpha^{(l-1)}\|_2 \leq  \prod_{k=1}^{l-1}\left( \sigma_1 + \frac{\rho}{\sqrt{m_{k}}}\right)  ~,
\end{align*}
which holds with the probability at least $1-\sum_{k=1}^{l-1}\frac{2}{m_{k}}$. Since $\phi$ is 1-Lipschitz, for layer $l$, we have
\begin{align*}
\left\|\phi\left( \frac{1}{\sqrt{m_{l}}} W^{(l)} \alpha^{(l-1)} \right) \right\|_2 - \| \phi(\mathbf{0}) \|_2 
&\leq \left\|\phi\left( \frac{1}{\sqrt{m_{l}}} W^{(l)} \alpha^{(l-1)} \right) - \phi(\mathbf{0}) \right\|_2\\
&\leq \left\|  \frac{1}{\sqrt{m_{l}}} W^{(l)} \alpha^{(l-1)} \right\|_2 ~,    
\end{align*}
so that
\begin{align*}
\|\alpha^{(l)}\|_2 & = \left\| \phi\left( \frac{1}{\sqrt{m_{l}}} W^{(l)} \alpha^{(l-1)} \right) \right\|_2 
\leq  \left\|  \frac{1}{\sqrt{m_{l}}} W^{(l)} \alpha^{(l-1)} \right\|_2 + \| \phi(\mathbf{0}) \|_2 \\
& \leq  \frac{1}{\sqrt{m_{l}}} \| W^{(l)} \|_2 \|\alpha^{(l-1)} \|_2   \\
& \overset{(a)}{\leq} \left( \sigma_1 + \frac{\rho}{\sqrt{m_{l}}}\right) \|\alpha^{(l-1)} \|_2   \\
& \overset{(b)}{\leq}  \prod_{k=1}^{l}\left( \sigma_1 + \frac{\rho}{\sqrt{m_{k}}}\right) ~,
\end{align*}
where (a) follows from Lemma~\ref{lemma:matrix} and (b) from the inductive step. Since we have used Lemma~\ref{lemma:matrix} $l$ times, after a union bound, our result would hold with probability at least $1-\sum_{k=1}^{l}\frac{2}{m_{k}}$.  This completes the proof. \qed 

Recall that in our setup, the layerwise outputs and pre-activations are respectively given by:
\begin{align}
\alpha^{(l)} = \phi\left(\tilde{\alpha}^{(l)} \right)~,~~~
\tilde{\alpha}^{(l)} := \frac{1}{\sqrt{m_{l}}} W^{(l)} \alpha^{(l-1)} ~.
\end{align}
\begin{lemm}
Consider any $l\in\{2,\dots,L\}$. Under Assumptions~\ref{asmp:act} and \ref{asmp:ginit}, for $\theta \in B_{\rho,\rho_1}^{\spec}(\theta_0)$, with probability at least $\left(1-\frac{2}{m_{l}}\right)$,
\begin{equation}
    \left\| \frac{\partial \alpha^{(l)}}{\partial \alpha^{(l-1)}} \right\|_2^2    \leq \left( \sigma_1 + \frac{\rho}{\sqrt{m_{l}}} \right)^2 = \beta_{l}^2~.
\end{equation}
\label{lemm:alpha_l_alpha_l-1}
\end{lemm}
\proof By definition, we have
\begin{align}
\left[ \frac{\partial \alpha^{(l)}}{\partial \alpha^{(l-1)}}  \right]_{i,j} = \frac{1}{\sqrt{m_{l}}} \phi'(\tilde{\alpha}^{(l)}_i) W_{ij}^{(l)}~.
\label{eq:d_alpha_d_alpha}
\end{align}
Since $\|A\|_2 = \sup_{\|\u\|_2=1} \| A \u\|_2$, so that $\| A\|_2^2 = \sup_{\|\u\|_2 = 1} \sum_i \langle \a_i, \u \rangle^2$, we have that for $2 \leq l \leq L$,
\begin{align*}
 \left\| \frac{\partial \alpha^{(l)}}{\partial \alpha^{(l-1)}} \right\|_2^2  & =  \sup_{\|\u\|_2=1} \frac{1}{m_{l}} \sum_{i=1}^{m_{l}} \left( \phi'(\tilde{\alpha}^{(l)}_i) \sum_{j=1}^{m_{l-1}}W_{ij}^{(l)} u_j \right)^2 \\
 & \overset{(a)}{\leq} \sup_{\|\u\|_2=1} \frac{1}{m_{l}} \| W^{(l)} \u \|_2^2 \\
 & = \frac{1}{m_{l}} \| W^{(l)} \|_2^2 \\
 & \overset{(b)}{\leq}\beta_{l}^2~,
\end{align*}
where (a) follows from $\phi$ being 1-Lipschitz by Assumption~\ref{asmp:act} and (b) from Lemma~\ref{lemma:matrix}. This completes the proof. \qed
\begin{lemm}\label{lemma:partial-layer}
Consider any $l\in[L]$. Denote the parameter vector $\w^{(l)}=\text{vec}\left(W^{(l)}\right)$. Under Assumption~\ref{asmp:act} and \ref{asmp:ginit}, for $\theta\in B_{\rho,\rho_{1}}^{\text{Spec}}\left(\theta_{0}\right)$ and $\beta_{k}=\sigma_{1}+\frac{\rho}{\sqrt{m_{k}}}$, with probability at least $1-\sum_{k=1}^{l-1}\frac{2}{m_{k}}$, 
\begin{align*}
    \left\|\frac{\partial\alpha^{(l)}}{\partial\w^{(l)}}\right\|_{2}^{2} \leq \frac{\prod_{k=1}^{l-1}\beta_{k}^{2}}{m_{l}}~.
\end{align*}
\end{lemm}
\proof Note that the parameter vector $\w^{(l)} = \text{vec}(W^{(l)})$ and can be indexed with $j\in[m_{1}]$ and 
$j'\in[d]$ when $l=1$, $j\in[m_{l}]$ and $j'\in[m_{l-1}]$ when $l\geq 2$. Then, we have
\begin{align}
 \left[ \frac{\partial \alpha^{(l)}}{\partial \w^{(l)}} \right]_{i,jj'} & = \left[ \frac{\partial \alpha^{(l)}}{\partial W^{(l)}} \right]_{i,jj'} = \frac{1}{\sqrt{m_{l}}} \phi'(\tilde{\alpha}^{(l)}_i) \alpha^{(l-1)}_{j'} \1_{[i=j]}~.
 \label{eq:d_alpha_d_w}
\end{align}
For $l\in\{2,\dots,L\}$, noting that $\frac{\partial \alpha^{(l)}}{\partial \w^{(l)}} \in \R^{m_{l} \times m_{l}m_{l-1}}$ and $\norm{V}_F=\norm{\vec(V)}_2$ for any matrix $V$, we have 
\begin{align*}
\left\| \frac{\partial \alpha^{(l)}}{\partial \w^{(l)}} \right\|_2^2  & = \sup_{\| V \|_F =1} \frac{1}{m_{l}} \sum_{i=1}^{m_{l}}  \left( \phi'(\tilde{\alpha}_i^{(l)} ) \sum_{j=1}^{m_{l}}\sum_{j'=1}^{m_{l-1}}\alpha^{(l-1)}_{j'} \1_{[i=j]} V_{jj'} \right)^2 \\
& \leq  \sup_{\| V \|_F =1} \frac{1}{m_{l}} \| V \alpha^{(l-1)} \|_2^2 \\
& \leq \frac{1}{m_{l}} \sup_{\| V \|_F =1} \| V \|_2^2 \| \alpha^{(l-1)} \|_2^2 \\
& \overset{(a)}{\leq} \frac{1}{m_{l}}  \| \alpha^{(l-1)} \|_2^2 \\
& \overset{(b)}{\leq} \frac{1}{m_{l}} \prod_{k=1}^{l-1}\beta_{k}^{2}~,
\end{align*}
where (a) follows from $\norm{V}_2^2\leq\norm{V}_F^2$ for any matrix $V$, and (b) from Lemma~\ref{lemma:layer}.

The $l=1$ case follows in a similar manner:
\begin{equation*}
\left\| \frac{\partial \alpha^{(1)}}{\partial \w^{(1)}} \right\|_2^2 \leq \frac{1}{m_{1}}  \| \alpha^{(0)} \|_2^2
=\frac{1}{m_{1}}\norm{\x}_2^2= \frac{1}{m_{1}}~,
\end{equation*} 
which satisfies the form for $l=1$. That completes the proof. \qed

Before getting into the Gaussian width analysis, we need two results which will be used in our main proofs.     
\begin{lemm}\label{lemm:minksum}
Let $\cA \in \R^{d_1 + d_2}$. For any $\a \in\cA$, let $\a_1 = \Pi_{\R^{d_1}}(\a)$, i.e., projection of $\a$ on the first $d_1$ coordinates, and let $\a_2 = \Pi_{\R^{d_2}}(\a)$, i.e., projection of $\a$ on the latter $d_2$ coordinates. Let $\cA_1 := \{\a_1 | \a \in\cA\}$ and $A_2 = \{\a_2 | \a \in \cA \}$. Then, 
\begin{align}
    w(\cA) \leq w(\cA_1) + w(\cA_2)~.
\end{align}
\end{lemm}
\proof Note that $\cA \subset \cA_1 + \cA_2$, where $+$ here denotes the Minkowski sum, i.e., all $\a' = \a_1 + \a_2$ for $\a_1 \in\cA_1, \a_2 \in \cA_2$. Then, 
\begin{align*}
w(A) & \leq w(\cA_1 + \cA_2) \\
& = \E_{\g_{d_1+d_2}}\left[ \sup_{\a_1 + \a_2 \in\cA_1 + \cA_2} \langle \a_1 + \a_2, \g_{d_1+d_2} \rangle \right] \\
& \leq \E_{\g_{d_1+d_2}}\left[ \sup_{\a_1 \in\cA_1 } \langle \a_1 , \g_{d_1} \rangle + \sup_{\a_2 \in\cA_2 } \langle \a_2 , \g_{d_2} \rangle\right] \\
& = \E_{\g_{d_1}}\left[ \sup_{\a_1 \in\cA_1 } \langle \a_1 , \g_{d_1} \rangle \right] + \E_{\g_{d_2}}\left[ \sup_{\a_2 \in\cA_2 } \langle \a_2 , \g_{d_2} \rangle\right] \\
& = w(\cA_1) + w(\cA_2)~.
\end{align*}
That completes the proof. \qed 
\begin{lemm}
\label{lemm:widthspectral}
Consider $\cA = \{\a \}$ with $\a \in \R^D$ such that each $\a$ can be written as $\a = B\v$ for some $B \in \cB \subset \R^{D \times d}, \v \in \cV \subset \R^d$. Then, there exists a constant $C$ such that 
\begin{align}
    w(\cA) \leq C\sup_{B \in \cB} \| B\|_2 w(\cV)~.
\end{align}
\end{lemm}
\proof
According to the definition of $\gamma_{2}$ function, we have
\begin{align*}
    \gamma_{2}\left(\cA,d\right)&=\inf_{\Gamma}\sup_{\a\in A}\sum_{k=0}^{\infty}2^{k/2}d\left(\a,\cA_{k}\right)\\
    \gamma_{2}\left(\cV,d\right)&=\inf_{\Lambda}\sup_{\v\in\cV}\sum_{k=0}^{\infty}2^{k/2}d\left(\v,\cV_{k}\right)
\end{align*}
where the infimum is taken with respect to all admissible sequences $\Gamma$ of $A$ and $\Lambda$ of $\cV$, and the distance $d$ is the $\ell_{2}$ distance for both spaces:
\begin{align*}
    d\left(\a_{1},\a_{2}\right)&:=\left\|\a_{1}-\a_{2}\right\|_{2}\\
    d\left(\v_{1},\v_{2}\right)&:=\left\|\v_{1}-\v_{2}\right\|_{2}
\end{align*}
For any fixed $\a\in\cA$, we can write it as $\a=B\v$. For any admissible sequence $\Lambda=\left\{\cV_{k}\right\}_{k=1}^{\infty}$ of $\cV$, we consider the following sequence:
\begin{align*}
    B\Lambda:=\left\{B\cV_{k}\right\}_{k=1}^{\infty}
\end{align*}
in which $B\cV_{k}:=\left\{B\v:\v\in\cV_{k}\right\}$. Since for each $k$, $B\cV_{k}$ is a subset of $\cA$, and $\left|B\cV_{k}\right|\leq\left|\cV_{k}\right|\leq2^{2^{k}}$, so $B\Lambda$ is a admissible sequence of $\cA$. Now in $\cV$, we walk from the original point $\v_{0}$ to $\v$ along the chain
\begin{align*}
    \v_{0}=\pi_{0}\left(\v\right)\rightarrow\pi_{1}\left(\v\right)\rightarrow\cdot\cdot\cdot\rightarrow\pi_{K}\left(\v\right)\rightarrow\cdot\cdot\cdot
\end{align*}
in which $\pi_{k}\left(\v\right)\in\cV_{k}$ is the best approximation to $\v$ in $\cV_{k}$, i.e.
\begin{align*}
    d\left(\v,\cV_{k}\right)=d\left(\v,\pi_{k}\left(\v\right)\right)
\end{align*}
then obviously $d\left(\a,B\cV_{k}\right)\leq d\left(\a,B\pi_{k}\left(\v\right)\right)$ and $B\pi_{k}\left(\v\right)\in B\cV_{k}$. Therefore,
\begin{align*}
    \sum_{k=0}^{\infty}2^{k/2}d(\a,B\cV_{k})&\leq\sum_{k=0}^{\infty}2^{k/2}d\left(B\v,B\pi_{k}\left(\v\right)\right)=\sum_{k=0}^{\infty}2^{k/2}\left\|B\v-B\pi_{k}\left(\v\right)\right\|_{2}\\
    &\leq\sum_{k=0}^{\infty}2^{k/2}\left\|B\right\|_{2}\left\|\v-\pi_{k}\left(\v\right)\right\|_{2}\leq\sup_{B\in\cB}\left\|B\right\|_{2}\sum_{k=0}^{\infty}2^{k/2}\left\|\v-\pi_{k}\left(\v\right)\right\|_{2}\\
    &=\sup_{B\in\cB}\left\|B\right\|_{2}\sum_{k=0}^{\infty}2^{k/2}d\left(\v,\pi_{k}\left(\v\right)\right)=\sup_{B\in\cB}\left\|B\right\|_{2}\sum_{k=0}^{\infty}2^{k/2}d\left(\v,\cV_{k}\right)\\
    &\leq\sup_{B\in\cB}\left\|B\right\|_{2}\sup_{\v\in\cV}\sum_{k=0}^{\infty}2^{k/2}d\left(\v,\cV_{k}\right)
\end{align*}
By taking supreme over $\a\in\cA$, we can get that
\begin{align*}
    \sup_{\a\in\cA}\sum_{k=0}^{\infty}2^{k/2}d(\a,B\cV_{k})
    \leq\sup_{B\in\cB}\left\|B\right\|_{2}\sup_{\v\in\cV}\sum_{k=0}^{\infty}2^{k/2}d\left(\v,\cV_{k}\right)
\end{align*}
Since all $B\Lambda$ are admissible sequences of $\cA$, then by taking infimum over all admissible sequences $\Lambda$ of $\cV$, we have
\begin{align*}
    \gamma_{2}(\cA,d)&=\inf_{\Gamma}\sup_{\a\in\cA}\sum_{k=0}^{\infty}2^{k/2}d(\a,\cA_{k})\leq\inf_{\Lambda}\sup_{\a\in\cA}\sum_{k=0}^{\infty}2^{k/2}d(\a,B\cV_{k})\\
    &\leq\inf_{\Lambda}\sup_{B\in\cB}\left\|B\right\|_{2}\sup_{\v\in\cV}\sum_{k=0}^{\infty}2^{k/2}d\left(\v,\cV_{k}\right)=\sup_{B\in\cB}\left\|B\right\|_{2}\inf_{\Lambda}\sup_{\v\in\cV}\sum_{k=0}^{\infty}2^{k/2}d\left(\v,\cV_{k}\right)\\
    &=\sup_{B\in\cB}\left\|B\right\|_{2}\gamma_{2}(\cV,d)
\end{align*}

For the mean-zero Gaussian process $X_{\a}=\left\langle\a,g\right\rangle$ on $\cA$, in which $g$ is a standard Gaussian vector in $\mathbb{R}^{D}$, we have
\begin{align*}
    \left\|X_{\a_{1}}-X_{\a_{2}}\right\|_{2}^{2}=\mathbb{E}_{g}\left|\left\langle\a_{1},g\right\rangle-\left\langle\a_{2},g\right\rangle\right|^{2}=\mathbb{E}_{g}\left\langle\a_{1}-\a_{2},g\right\rangle^{2}=\left\|\a_{1}-\a_{2}\right\|_{2}^{2}
\end{align*}
so $d\left(\a_{1},\a_{2}\right)=\left\|\a_{1}-\a_{2}\right\|_{2}=\left\|X_{\a_{1}}-X_{\a_{2}}\right\|_{2}$ is the canonical distance on $\cA$. Similarly, for the mean-zero Gaussian process $Y_{\v}=\left\langle\v,h\right\rangle$ on $\cV$, in which $h$ is a standard Gaussian vector in $\mathbb{R}^{d}$,
\begin{align*}
d\left(\v_{1},\v_{2}\right)=\left\|\v_{1}-\v_{2}\right\|_{2}=\left\|Y_{\v_{1}}-Y_{\v_{2}}\right\|_{2}
\end{align*}
is also the canonical distance on $\cV$. Then by Talagrand’s majorizing measure theorem, there are absolute constants $c, C$ such that
\begin{align*}
    w\left(\cA\right)&=\mathbb{E}_{g}\sup_{\a\in\cA}X_{\a}\leq c\gamma_{2}\left(\cA,d\right)\leq c\sup_{B\in\cB}\left\|B\right\|_{2}\gamma_{2}(\cV,d)\\
    &\leq C\sup_{B\in\cB}\left\|B\right\|_{2}\mathbb{E}_{h}\sup_{\v\in\cV}Y_{\v}=C\sup_{B\in\cB}\left\|B\right\|_{2}w\left(\cV\right)
\end{align*}
so the inequality holds. \qed 

We now restate and prove the main result for FFNs.
\theofnngradwidth*
\proof The loss at any input $(\x,y)$ is given by $\ell(y,f(\theta;\x))$ with $\theta\in B_{\rho,\rho_{1}}^{\text{Spec}}\left(\theta_{0}\right)$. With $\hat{y} = f(\theta;\x)$, let $\ell' := \frac{d \ell(y,\hat{y})}{d \hat{y}}$. Then, the gradient of the loss 
\begin{equation}
    \nabla_{\theta} \ell(y, f(\theta;\x)) = \ell' \nabla_{\theta} f(\theta,\x)~.
\end{equation}
With $\ell' = O(1)$, we focus our analysis on the Gaussian width of the gradient of the predictor $\nabla_{\theta} f(\theta,\x)$, and the Gaussian width of the loss gradient set will be bounded by a constant times the Gaussian width of predictor gradient set. 

Recall that
\begin{equation}
f(\theta;\x) = \v^\top \phi(\frac{1}{\sqrt{m_{L}}}W^{(L)} \phi(\cdots \phi(\frac{1}{\sqrt{m_{1}}}  W^{(1)} \x))))~,
\end{equation}
where
\begin{align}
\theta = (\vec(W^{(1)})^\top,\ldots,\vec(W^{(L)})^\top, \v^\top )^\top~.
\end{align}
For convenience, we write the model in terms of the layerwise outputs or features as:
\begin{align}
\alpha^{(0)}(\x) & = \x~, \\
\alpha^{(l)} & = \phi\left( \frac{1}{\sqrt{m_{l}}} W^{(l)} \alpha^{(l-1)}(\x) \right) \\
f(\theta;\x) & = \v^\top \alpha^{(L)}(\x)~.
\end{align}
By Lemma~\ref{lemm:minksum}, we bound the Gaussian width of the overall gradient by the sum of the width of the gradients of the last layer parameters $\v \in \R^m$ and that of all the intermediate parameters $W^{(l)}, \ell \in [L]$. 

Starting with the last layer parameter $\v \in \R^m$, the gradient is in $\R^m$ and we have 
\begin{align}
\frac{\partial f}{\partial \v} = \alpha^{(L)}(\x)~,
\end{align}
which is the output of the last layer or the so-called featurizer, i.e., $h^{(L)}(W,\x) = \alpha^{(L)}(\x)$. Then, the Gaussian width is simply $w(A^{(L)})$. 

For any hidden layer parameter $\w^{(l)} \in \R^{m_{l}m_{l-1}}$ with $\w^{(l)} = \vec(W^{(l)})$, the gradient is in $\R^{m_{l}m_{l-1}}$ and we have
\begin{align}
\frac{\partial f}{\partial \w^{(l)}} & = \frac{\partial \alpha^{(l)}}{\partial \w^{(l)}} \frac{\partial f}{\partial \alpha^{(l)}} \notag\\
& = \frac{\partial \alpha^{(l)}}{\partial \w^{(l)}} \left( \prod_{l'=l+1}^L \frac{\partial \alpha^{(l')}}{\partial \alpha^{(l'-1)}} \right) \frac{\partial f}{\partial  \alpha^{(L)}} ~\label{eq:paramexpan}
\end{align}
Define $\mathcal{Z}^{(l)}=\left\{Z^{(l)}\in\mathbb{R}^{m_{l}\times m_{l}m_{l-1}}:\left\|Z^{(l)}\right\|_{2}\leq\frac{\prod_{k=1}^{l-1}\beta_{k}}{\sqrt{m_{l}}}\right\}$, then according to Lemma \ref{lemma:partial-layer}, $\frac{\partial \alpha^{(l)}}{\partial \w^{(l)}}\in\mathcal{Z}^{(l)}$ with probability $1-\sum_{k=1}^{l-1}\frac{2}{m_{k}}$. Similarly, define $\mathcal{B}^{(l)}=\left\{B^{(l)}\in\mathbb{R}^{m_{l}\times m_{l-1}}:\left\|B^{(l)}\right\|_{2}\leq\beta_{l}\right\}$, then according to Lemma \ref{lemm:alpha_l_alpha_l-1}, $\frac{\partial \alpha^{(l)}}{\partial \alpha^{(l-1)}}\in\mathcal{B}^{(l)}$ with probability $1-\frac{2}{m_{l}}$. Besides, define $\mathcal{V}$ the set of $\v$, then according to \eqref{eq:specball_app}, $\v\in\mathbb{R}^{m_{L}}$ and $\left\|\v\right\|_{2}\leq1+\rho_{1}$, so $w\left(\mathcal{V}\right)\leq c\sup_{\v\in\mathcal{V}}\left\|\v\right\|_{2}\sqrt{m_{L}}\leq c(1+\rho_{1})\sqrt{m_{L}}$.

Therefore, with $\mathcal{H}^{(l)}=\left\{\frac{\partial f}{\partial \w^{(l)}}\mid W \in B_{\rho}^{\spec}(\theta_0) \right\}$, according to \eqref{eq:paramexpan}, we have
\begin{align*}
w(\mathcal{H}^{(l)} ) 
& \overset{(a)}{\leq} C\sup_{Z^{(l)}\in\mathcal{Z}^{(l)}} \| Z^{(l)} \|_2 \prod_{l'=l+1}^L \sup_{B^{(l')}\in\mathcal{B}^{l'}}\| B^{(l')} \|_2 ~ w(\mathcal{V}) \\
& \overset{(b)}{\leq} C\frac{\prod_{k=1}^{l-1}\beta_{k}}{\sqrt{m_{l}}} \prod_{k=l+1}^{L}\beta_{k}\cdot c\left(1+\rho_{1}\right)\sqrt{m_{L}} \\
& = C_{0}\sqrt{\frac{m_{L}}{m_{l}}}(1+\rho_{1})\prod_{k\not=l}\beta_{k}~.
\end{align*}
where (a) follows from Lemma \ref{lemm:widthspectral} and (b) from the definition of $\mathcal{Z}^{(l)}$ and $\mathcal{B}^{(l)}$. 

Since 
\begin{align*}
    \theta = (\vec(W^{(1)})^\top,\ldots,\vec(W^{(L)})^\top, \v^\top )^\top=\left(\w^{(1)}{}^{\top},...,\w^{(L)}{}^{\top},\v^{\top}\right)^{\top}
\end{align*} 
According to Lemma \ref{lemm:minksum}, the Gaussian width of the individual predictor gradient set is bounded by
\begin{align*}
\sum_{l=1}^{L}w(\mathcal{H}^{(l)})+w(A^{(L)})&\leq\sum_{l=1}^{L}C_{0}\sqrt{\frac{m_{L}}{m_{l}}}(1+\rho_{1})\prod_{k\not=l}\beta_{k}+w(A^{(L)})\\
&=C_{0}(1+\rho_{1})\sqrt{m_{L}}\prod_{l=1}^{L}\beta_{l}\sum_{l=1}^{L}\frac{1}{\beta_{l}\sqrt{m_{l}}}+w(A^{(L)})
\end{align*}
then there exists constant $c_{1}, c_{2}$ such that 
\begin{align*}
    w(\Xi^{\text{ffn}})\leq c_{1}w(A^{(L)})+c_{2}(1+\rho_{1})\sqrt{m_{L}}\prod_{l=1}^{L}\beta_{l}\sum_{l=1}^{L}\frac{1}{\beta_{l}\sqrt{m_{l}}}
\end{align*}

Since the spectral norm bound for each layer except the last layer holds with probability $(1-\frac{2}{m_{l}})$ according to Lemma \ref{lemma:matrix}, and we use it $L$ times in the proof, the Gaussian width bound holds with probability $(1-\sum_{l=1}^{L}\frac{2}{m_{l}})$. 
\qed 

\subsection{Gaussian Width bounds for Residual Networks (ResNets)}
\label{app:resnet}
We now consider the setting where $f$ is a residual network (ResNet) with depth $L$
 and widths $m_l, l \in [L] := \{1,\ldots,L\}$ given by
\begin{equation}
\begin{split}
    \alpha^{(0)}(\x) & = \x~, \\
     \alpha^{(l)}(\x) & = \alpha^{(l-1)}(\x) + \phi \left( \frac{1}{L \sqrt{m_{l}}} W^{(l)} \alpha^{(l-1)}(\x) \right)~,~~~l=1,\ldots,L~,\\
    f(\theta;\x)  = \alpha^{(L+1)}(\x) & = \v^\top \alpha^{(L)}(\x)~,
    \end{split}
\label{eq:DNN}    
\end{equation}
where $W^{(l)} \in \R^{m_l \times m_{l-1}}, l \in [L]$ are layer-wise weight matrices, $\v\in\R^{m_{L}}$ is the last layer vector, $\phi(\cdot )$ is the smooth (pointwise) activation function, and the total set of parameters 
\begin{align}
\theta := (\vec(W^{(1)})^\top,\ldots,\vec(W^{(L)})^\top, \v^\top )^\top \in \R^{\sum_{k=1}^L m_k m_{k-1}+m_{L}}~,
\label{eq:theta_def_2}
\end{align}
with $m_0=d$.

We start by bounding the norm of the output $\alpha^{(l)}$ of layer $l$. 
\begin{lemm}
Consider any $l\in[L]$ and let $\beta := \sigma_1 + \frac{\rho}{\sqrt{m}}$. 
Under Assumptions~\ref{asmp:act} and \ref{asmp:ginit}, for $\theta \in B_{\rho,\rho_1}^{\spec}(\theta_0)$, with probability at least $\left(1 - \sum_{k=1}^{l}\frac{2}{m_{k}} \right)$, we have
\begin{align}
\| \alpha^{(l)}\|_2 \leq \prod_{k=1}^{l}\left(1 + \frac{\beta_{k}}{L} \right) ~.
\end{align}
\label{lemm:outl2_res}
\end{lemm}
\proof Following~\cite{allen-zhu_convergence_2019,CL-LZ-MB:20}, we prove the result by recursion. First, recall that since $\| \x\|_2^2 = 1$, we have $\| \alpha^{(0)}\|_2 = 1$. Then, $\phi$ is 1-Lipschitz,
\begin{align*}
\left\|\phi\left( \frac{1}{L\sqrt{m_{1}}} W^{(1)} \alpha^{(0)} \right) \right\|_2 - \| \phi(\mathbf{0}) \|_2 
\leq \left\|\phi\left( \frac{1}{L\sqrt{m_{1}}} W^{(1)} \alpha^{(0)} \right) - \phi(\mathbf{0}) \right\|_2 \leq \left\|  \frac{1}{L\sqrt{m_{1}}} W^{(1)} \alpha^{(0)} \right\|_2 ~,
\end{align*}
so that, using $\phi(0)=0$, we have 
\begin{align*}
\| \alpha^{(1)}\|_2 & \leq \| \alpha^{(0)} \|_2 + \left\| \phi\left( \frac{1}{L\sqrt{m_{1}}} W^{(1)} \alpha^{(0)} \right) \right\|_2 \\
& \leq 1 +  \left\|  \frac{1}{L\sqrt{m_{1}}} W^{(1)} \alpha^{(0)} \right\|_2 \\
&\leq 1 +  \frac{1}{L\sqrt{m_{1}}} \| W^{(1)} \|_2 \|\alpha^{(0)} \|_2  \\
& \overset{(a)}{\leq}  1 + \frac{\beta_{1}}{L}~,
\end{align*}
where (a) follows from Lemma~\ref{lemma:matrix} which holds with probability at least $1-\frac{2}{m_{1}}$. 

For the inductive step, we assume that for some $(l-1)$, we have 
\begin{align*}
\| \alpha^{(l-1)}\|_2 \leq \prod_{k=1}^{l-1}\left( 1 + \frac{\beta_{k}}{L} \right)~, 
\end{align*}
which holds with the probability at least $1-\sum_{k=1}^{l-1}\frac{2}{m_{k}}$. Since $\phi$ is 1-Lipschitz, for layer $l$, we have
\begin{align*}
\left\|\phi\left( \frac{1}{L\sqrt{m_{l}}} W^{(l)} \alpha^{(l-1)} \right) \right\|_2 - \| \phi(\mathbf{0}) \|_2 
&\leq \left\|\phi\left( \frac{1}{L\sqrt{m_{l}}} W^{(l)} \alpha^{(l-1)} \right) - \phi(\mathbf{0}) \right\|_2
\leq \left\|  \frac{1}{L\sqrt{m_{l}}} W^{(l)} \alpha^{(l-1)} \right\|_2 ~,    
\end{align*}
so that, using $\phi(0)=0$, we have
\begin{align*}
\|\alpha^{(l)}\|_2 & \leq \| \alpha^{(l-1)} \|_2 + \left\| \phi\left( \frac{1}{L \sqrt{m_{l}}} W^{(l)} \alpha^{(l-1)} \right) \right\|_2 \\
& \leq  \| \alpha^{(l-1)} \|_2 + \left\|  \frac{1}{L\sqrt{m_{l}}} W^{(l)} \alpha^{(l-1)} \right\|_2  \\
& \leq \| \alpha^{(l-1)} \|_2 + \frac{1}{L \sqrt{m_{l}}} \| W^{(l)} \|_2 \|\alpha^{(l-1)} \|_2   \\
& \overset{(a)}{\leq} \| \alpha^{(l-1)} \|_2 + \frac{\beta_{l}}{L} \|\alpha^{(l-1)} \|_2   \\
& = \left( 1 + \frac{\beta_{l}}{L} \right) \| \alpha^{((l-1)} \|_2 \\
& \overset{(b)}{\leq} \prod_{k=1}^{l}\left( 1 + \frac{\beta_{k}}{L} \right)~,
\end{align*}
where (a) follows from Lemma~\ref{lemma:matrix} and (b) follows from the inductive step. Since we have used Lemma~\ref{lemma:matrix} $l$ times, after a union bound, our result would hold with probability at least $1-\sum_{k=1}^{l}\frac{2}{m_{k}}$.  That completes the proof. \qed 

Recall that in our setup, the layerwise outputs and pre-activations are respectively given by:
\begin{align}
\alpha^{(l)} = \alpha^{(l-1)} + \phi\left(\tilde{\alpha}^{(l)} \right)~,~~~
\tilde{\alpha}^{(l)} := \frac{1}{L \sqrt{m_{l}}} W^{(l)} \alpha^{(l-1)} ~.
\end{align}
\begin{lemm}
Consider any $l\in \{2,\dots,L\}$ and let $\beta_{l} := \sigma_1 + \frac{\rho}{\sqrt{m_{l}}}$. Under Assumptions~\ref{asmp:act} and \ref{asmp:ginit}, for $\theta \in B_{\rho,\rho_1}^{\spec}(\theta_0)$, with probability at least $\left(1-\frac{2}{m_{l}}\right)$,
\begin{equation}
    \left\| \frac{\partial \alpha^{(l)}}{\partial \alpha^{(l-1)}} \right\|_2    \leq  1 + \frac{\beta_{l}}{L}  ~.
\end{equation}
\label{lemm:alpha_l_alpha_l-1_res}
\end{lemm}
\proof By definition, we have
\begin{align}
\left[ \frac{\partial \alpha^{(l)}}{\partial \alpha^{(l-1)}}  \right]_{i,j} = \1_{[i=j]} + \frac{1}{L\sqrt{m_{l}}} \phi'(\tilde{\alpha}^{(l)}_i) W_{ij}^{(l)}~.
\label{eq:d_alpha_d_alpha_res}
\end{align}
so that with $C^{(l)}=[c^{(l)}_{ij}], c^{(l)}_{ij} = \frac{1}{L\sqrt{m}} \phi'(\tilde{\alpha}^{(l)}_i) W_{ij}^{(l)}$, we have 
\begin{align}
  \left\| \frac{\partial \alpha^{(l)}}{\partial \alpha^{(l-1)}} \right\|_2 = \left\| \I + C^{(l)} \right\|_2  \leq 1 + \| C^{(l)} \|_2~.  
\label{eq:spec_d_alpha_d_alpha_res}
\end{align}
Since $\|C^{(l)}\|_2 = \sup_{\|\u\|_2=1} \| C^{(l)} \u\|_2$, so that $\| C^{(l)}\|_2^2 = \sup_{\|\u\|_2 = 1} \sum_i \langle \c^{(l)}_i, \u \rangle^2$, we have that for $2 \leq l \leq L$,
\begin{align*}
 \left\| C^{(l)} \right\|_2^2  & =  \sup_{\|\u\|_2=1} \frac{1}{L^2 m_{l}} \sum_{i=1}^{m_{l}} \left( \phi'(\tilde{\alpha}^{(l)}_i) \sum_{j=1}^{m_{l-1}}W_{ij}^{(l)} u_j \right)^2 \\
 & \overset{(a)}{\leq} \sup_{\|\u\|_2=1} \frac{1}{L^2 m_{l}} \| W^{(l)} \u \|_2^2 \\
 & = \frac{1}{L^2 m_{l}} \| W^{(l)} \|_2^2 \\
 & \overset{(b)}{\leq} \left(\frac{\beta_{l}}{L} \right)^2~,
\end{align*}
where (a) follows from $\phi$ being 1-Lipschitz by Assumption~\ref{asmp:act} and (b) from Lemma~\ref{lemma:matrix}. Putting the bound back in \eqref{eq:spec_d_alpha_d_alpha_res} completes the proof. \qed 
\begin{lemm}
Consider any $l\in[L]$. 
Under Assumptions~\ref{asmp:act} and \ref{asmp:ginit}, for $\theta \in B_{\rho,\rho_1}^{\spec}(\theta_0)$, with probability at least $\left(1-\sum_{k=1}^{l}\frac{2}{m_{k}}\right)$,
\begin{equation}
\begin{split}
\left\| \frac{\partial \alpha^{(l)}}{\partial \w^{(l)}} \right\|_2  
&\leq \frac{1}{L\sqrt{m_{l}}} \prod_{k=1}^{l-1}\left( 1+ \frac{\beta_{k}}{L} \right) ~.
\end{split}
\end{equation}     
\label{lem:alpha_W_res}
\end{lemm}
\proof Note that the parameter vector $\w^{(l)} = \text{vec}(W^{(l)})$ and can be indexed with $j\in[m_{1}]$ and 
$j'\in[d]$ when $l=1$, $j\in[m_{l}]$ and $j'\in[m_{l-1}]$ when $l\geq 2$. Then, we have
\begin{align}
 \left[ \frac{\partial \alpha^{(l)}}{\partial \w^{(l)}} \right]_{i,jj'} & = \left[ \frac{\partial \alpha^{(l)}}{\partial W^{(l)}} \right]_{i,jj'} = \frac{1}{L\sqrt{m_{l}}} \phi'(\tilde{\alpha}^{(l)}_i) \alpha^{(l-1)}_{j'} \1_{[i=j]}~.
 \label{eq:d_alpha_d_w_res}
\end{align}
For $l\in\{2,\dots,L\}$, noting that $\frac{\partial \alpha^{(l)}}{\partial \w^{(l)}} \in \R^{m_{l} \times m_{l}m_{k-1}}$ and $\norm{V}_F=\norm{\vec(V)}_2$ for any matrix $V$, we have 
\begin{align*}
\left\| \frac{\partial \alpha^{(l)}}{\partial \w^{(l)}} \right\|_2^2  & = \sup_{\| V \|_F =1} \frac{1}{L^2 m_{l}} \sum_{i=1}^{m_{l}}  \left( \phi'(\tilde{\alpha}_i^{(l)} ) \sum_{j=1}^{m_{l}}\sum_{j'=1}^{m_{l-1}}\alpha^{(l-1)}_{j'} \1_{[i=j]} V_{jj'} \right)^2 \\
& \leq  \sup_{\| V \|_F =1} \frac{1}{L^2 m_{l}} \| V \alpha^{(l-1)} \|_2^2 \\
& \leq \frac{1}{L^2 m_{l}} \sup_{\| V \|_F =1} \| V \|_2^2 \| \alpha^{(l-1)} \|_2^2 \\
& \overset{(a)}{\leq} \frac{1}{L^2 m_{l}}  \| \alpha^{(l-1)} \|_2^2 \\
& \overset{(b)}{\leq} \frac{1}{L^2 m_{l}} \prod_{k=1}^{l-1}\left( 1 + \frac{\beta_{k}}{L} \right)^{2}
\end{align*}
where (a) follows from $\norm{V}_2^2\leq\norm{V}_F^2$ for any matrix $V$, and (b) from Lemma~\ref{lemm:outl2_res}.

The $l=1$ case follows in a similar manner:
\begin{equation*}
\left\| \frac{\partial \alpha^{(1)}}{\partial \w^{(1)}} \right\|_2^2 \leq \frac{1}{L^2 m_{1}}  \| \alpha^{(0)} \|_2^2
=\frac{1}{L^2 m_{1}}\norm{\x}_2^2 = \frac{1}{L^2 m_{1}}~,
\end{equation*} 
which satisfies the form for $l=1$. That completes the proof. \qed
We now restate and prove the main result for ResNets. 
\theorngradwidth*
\proof The loss at any input $(\x,y)$ is given by $\ell(y,f(\theta;\x))$ with $\theta\in B_{\rho,\rho_{1}}^{\text{Spec}}\left(\theta_{0}\right)$. With $\hat{y} = f(\theta;\x)$, let $\ell' := \frac{d \ell(y,\hat{y})}{d \hat{y}}$. Then, the gradient of the loss 
\begin{equation}
    \nabla_{\theta} \ell(y, f(\theta;\x)) = \ell' \nabla_{\theta} f(\theta,\x)~.
\end{equation}
With $\ell' = O(1)$, 
we focus our analysis on the Gaussian width of the gradient of the predictor $\nabla_{\theta} f(\theta,\x)$, and the Gaussian width of the loss gradient set will be bounded by a constant times the Gaussian width of predictor gradient set.

Recall that
 \begin{equation}
 f(\theta;x) = \v^\top \phi(\frac{1}{\sqrt{m_{L}}}W^{(L)} \phi(\cdots \phi(\frac{1}{\sqrt{m_{1}}}  W^{(1)} \x))))~,
 \end{equation}
 where
 \begin{align}
 \theta = (\vec(W^{(1)})^\top,\ldots,\vec(W^{(L)})^\top, \v^\top )^\top~.
 \end{align}
For convenience, we write the model in terms of the layerwise outputs or features as:
\begin{align}
\alpha^{(0)}(\x) & = \x~, \\
\alpha^{(l)} & = \alpha^{(l-1)} + \phi\left( \frac{1}{L\sqrt{m_{l}}} W^{(l)} \alpha^{(l-1)}(\x) \right) \\
f(\theta;\x) & = \v^\top \alpha^{(L)}(\x)~.
\end{align}
By Lemma~\ref{lemm:minksum}, we bound the Gaussian width of the overall gradient by the sum of the width of the gradients of the last layer parameters $\v \in \R^{m_{L}}$ and that of all the intermediate parameters $W^{(l)}, \ell \in [L]$. 

Starting with the last layer parameter $\v \in \R^m$, the gradient is in $\R^m$ and we have 
\begin{align}
\frac{\partial f}{\partial \v} = \alpha^{(L)}(\x)~,
\end{align}
which is the output of the last layer or the so-called featurizer. Then, the Gaussian width of the average set is simply $w(A^{(L)})$. 

For any hidden layer parameter $\w^{(l)} \in \R^{m_{l}m_{l-1}}$ with $\w^{(l)} = \vec(W^{(l)})$, the gradient is in $\R^{m_{l}m_{l-1}}$ and we have
\begin{align}
\frac{\partial f}{\partial \w^{(l)}} & = \frac{\partial \alpha^{(l)}}{\partial \w^{(l)}} \frac{\partial f}{\partial \alpha^{(l)}} \notag\\
& = \frac{\partial \alpha^{(l)}}{\partial \w^{(l)}} \left( \prod_{l'=l+1}^L \frac{\partial \alpha^{(l')}}{\partial \alpha^{(l'-1)}} \right) \frac{\partial f}{\partial \alpha^{(l)}}~
\label{eq:gradexpand_res}
\end{align}
Define $\mathcal{Z}^{(l)}=\left\{Z^{(l)}\in\mathbb{R}^{m_{l}\times m_{l}m_{l-1}}:\left\|Z^{(l)}\right\|_{2}\leq\frac{1}{L\sqrt{m_{l}}}\prod_{k=1}^{l-1}\left(1+\frac{\beta_{k}}{L}\right)\right\}$, then according to Lemma \ref{lem:alpha_W_res}, $\frac{\partial \alpha^{(l)}}{\partial \w^{(l)}}\in\mathcal{Z}^{(l)}$ with probability $1-\sum_{k=1}^{l-1}\frac{2}{m_{k}}$. Similarly, define $\mathcal{B}^{(l)}=\left\{B^{(l)}\in\mathbb{R}^{m_{l}\times m_{l-1}}:\left\|B^{(l)}\right\|_{2}\leq1+\frac{\beta_{l}}{L}\right\}$, then according to Lemma \ref{lemm:alpha_l_alpha_l-1_res}, $\frac{\partial \alpha^{(l)}}{\partial \alpha^{(l-1)}}\in\mathcal{B}^{(l)}$ with probability $1-\frac{2}{m_{l}}$. Besides, define $\mathcal{V}$ the set of $\v$, then according to \eqref{eq:specball_app}, $\v\in\mathbb{R}^{m_{L}}$ and $\left\|\v\right\|_{2}\leq1+\rho_{1}$, so $w\left(\mathcal{V}\right)\leq c\sup_{\v\in\mathcal{V}}\left\|\v\right\|_{2}\sqrt{m_{L}}\leq c(1+\rho_{1})\sqrt{m_{L}}$.

Therefore, with $\mathcal{H}^{(l)}=\left\{\frac{\partial f}{\partial \w^{(l)}}\mid W \in B_{\rho}^{\spec}(\theta_0) \right\}$, according to \eqref{eq:gradexpand_res}, we have
\begin{align*}
w(\mathcal{H}^{(l)} ) 
& \overset{(a)}{\leq} C\sup_{Z^{(l)}\in\mathcal{Z}^{(l)}} \| Z^{(l)} \|_2 \prod_{l'=l+1}^L \sup_{B^{(l)}\in\mathcal{B}^{(l)}}\| B^{(l)} \|_2 ~ w(\mathcal{V}) \\
& \overset{(b)}{\leq} C\frac{1}{L\sqrt{m_{l}}}\prod_{k=1}^{l-1}\left(1+\frac{\beta_{k}}{L}\right) \prod_{k=l+1}^{L}\left(1+\frac{\beta_{k}}{L}\right)\cdot c\left(1+\rho_{1}\right)\sqrt{m_{L}} \\
& = C_{0}\frac{1+\rho_{1}}{L}\sqrt{\frac{m_{L}}{m_{l}}}\prod_{k\not=l}\left(1+\frac{\beta_{k}}{L}\right)~
\end{align*}
where (a) follows from Lemma \ref{lemm:widthspectral}, (b) from the definition of $\mathcal{Z}^{(l)}$ and $\mathcal{B}^{(l)}$.
Since 
\begin{align*}
    \theta = (\vec(W^{(1)})^\top,\ldots,\vec(W^{(L)})^\top, \v^\top )^\top=\left(\w^{(1)}{}^{\top},...,\w^{(L)}{}^{\top},\v^{\top}\right)^{\top}
\end{align*} 
According to Lemma \ref{lemm:minksum}, the Gaussian width of the average predictor gradient set is bounded by 
\begin{align*}
    \sum_{l=1}^{L}w(\mathcal{H}^{(l)})+w(A^{(L)})&\leq \sum_{l=1}^{L}C_{0}\frac{1+\rho_{1}}{L}\sqrt{\frac{m_{L}}{m_{l}}}\prod_{k\not=l}\left(1+\frac{\beta_{k}}{L}\right)+w(A^{(L)})\\
    &=C_{0}\frac{1+\rho_{1}}{L}\sqrt{m_{L}}\prod_{l=1}^{L}\left(1+\frac{\beta_{l}}{L}\right)\sum_{l=1}^{L}\frac{1}{\left(1+\frac{\beta_{l}}{L}\right)\sqrt{m_{l}}}+w(A^{(L)})
\end{align*}
then there exists constant $c_{1}, c_{2}$ such that 
\begin{align*}
    w(\Xi^{\text{rn}})\leq c_{1}w(A^{(L)})+c_{2}\frac{1+\rho_{1}}{L}\sqrt{m_{L}}\prod_{l=1}^{L}\left(1+\frac{\beta_{l}}{L}\right)\sum_{l=1}^{L}\frac{1}{\left(1+\frac{\beta_{l}}{L}\right)\sqrt{m_{l}}}
\end{align*}

Since the spectral norm bound for each layer except the last layer holds with probability $(1-\frac{2}{m_{l}})$ according to Lemma \ref{lemma:matrix}, and we use it $L$ times in the proof, the Gaussian width bound holds with probability $(1-\sum_{l=1}^{L}\frac{2}{m_{l}})$. 
\qed


\end{document}